\documentclass{article}

\usepackage[
  margin=0.75in,               
  includefoot, heightrounded,
  marginparwidth=3.5cm,     
  marginparsep=2mm          
]{geometry}
\usepackage{times} 

\usepackage[parfill]{parskip}

\newcommand{\stopgrad}{\textbf{sg}}

\newcommand{\hidden}{\mathbf{h}}

\usepackage[utf8]{inputenc}  
\usepackage[T1]{fontenc}    
\usepackage{url}           
\usepackage{booktabs}     
\usepackage{amsfonts,bm}     
\usepackage{nicefrac}      
\usepackage{microtype}       
\usepackage{tikz}
\usepackage{svg}
\usetikzlibrary{fit,positioning,calc,shapes,arrows.meta}
\usepackage{subcaption}
\usepackage{soul}
\usepackage{csquotes}

\usepackage{natbib,hyperref}

\usepackage{color,xcolor}
\definecolor{darkblue}{rgb}{0, 0, 0.5}
\definecolor{darkgreen}{rgb}{0, 0.6, 0}
\definecolor{darkred}{rgb}{0.6, 0, 0}

\definecolor{gsoftcolor}{HTML}{0B3D91}
\definecolor{grescolor}{HTML}{A30000}
\NewDocumentCommand{\gsoft}{o}{{\color{gsoftcolor}g\IfValueT{#1}{^{#1}}_{\mathrm{KL}}}}
\NewDocumentCommand{\gres}{o}{{\color{grescolor}g\IfValueT{#1}{^{#1}}_{\text{res}}}}
\hypersetup{colorlinks=true, citecolor=darkblue, linkcolor=darkblue, urlcolor=darkblue}

\usepackage[colorinlistoftodos]{todonotes} 

\usepackage{amsmath}
\usepackage{amssymb}
\usepackage{mathtools} 
\usepackage{amsthm}

\usepackage{amssymb}
\usepackage{xcolor}
\usepackage{wrapfig}

\usepackage{enumitem}

\setlist{nosep,leftmargin=0.3in} 
\usepackage{multirow}
\usepackage[noend]{algpseudocode}
\usepackage{algorithm}

\usepackage{graphicx}  
\usepackage{xurl}

\usepackage[capitalize,noabbrev,nameinlink]{cleveref}

\usepackage{mdframed}
\mdfsetup{
linecolor=black!40,
linewidth=0.5pt,
innerleftmargin = 2pt,
innertopmargin = 4pt,
innerrightmargin= 2pt,
innerbottommargin = 4pt,
}

\usepackage{listings}
\definecolor{codegreen}{rgb}{0,0.6,0}
\definecolor{codegray}{rgb}{0.5,0.5,0.5}
\definecolor{codepurple}{rgb}{0.58,0,0.82}
\definecolor{backcolour}{rgb}{0.95,0.95,0.92}

\definecolor{lightgray}{gray}{0.9}

\usepackage{xcolor}
\usepackage[most]{tcolorbox}
\tcbuselibrary{listings,skins,breakable}

\definecolor{codebg}{HTML}{FCFCFD}
\definecolor{codeborder}{HTML}{D7D9DE}
\definecolor{codetext}{HTML}{000000}
\definecolor{codecomment}{HTML}{8D929C}
\definecolor{codeadd}{HTML}{3F5FD7}
\newcommand{\codesize}{\fontsize{7}{9.6}\selectfont}    
\newcommand{\hlcode}[1]{%
    \begingroup\setlength{\fboxsep}{1.2pt}%
    \frenchspacing\colorbox{codehl}{\ttfamily\codesize\color{codeadd}\strut #1}\endgroup%
}

\lstdefinestyle{minimalcode}{
    language=Python,
    basicstyle=\ttfamily\footnotesize\color{codetext},
    keywordstyle=\color{codetext},       
    identifierstyle=\color{codetext},
    commentstyle=\itshape\color{codecomment},
    stringstyle=\color{codetext},
    showstringspaces=false,
    keepspaces=true,
    columns=fullflexible,
    breaklines=true,
    tabsize=4,
    numbers=none,
    frame=none,
    upquote=true,
    moredelim=**[is][\color{codeadd}]{(*@}{@*)},
    escapeinside={(*!}{!*)},
}

\newtcblisting{codebubble}[1][]{%
    enhanced,
    listing only,
    breakable,
    listing engine=listings,
    listing options={style=minimalcode},
    colback=codebg,
    colframe=codeborder,
    boxrule=0.5pt,
    arc=4mm,
    outer arc=4mm,
    left=3mm,
    right=3mm,
    top=2mm,
    bottom=2mm,
    #1
}

\newlength{\coverbodyh}                
\newlength{\coverdivbot}               
\tikzset{coverdots/.style={black!35, line width=0.7pt, dash pattern=on 1pt off 2pt}}

\tikzset{
    lens/.style={codeadd!55, line width=0.6pt},
    lensbox/.style={lens, draw, rounded corners=2.5pt, inner sep=0pt},
}
\newcommand{\codemark}[1]{\tikz[remember picture, overlay]\coordinate (#1) at (0,0);}

\newcommand{\coverhead}[2][0pt]{%
    \hspace*{\dimexpr#1*2\relax}{\small\bfseries #2}\\[9pt]%
}
\newcommand{\coverdivider}{%
    \tikz[baseline=0pt]\draw[coverdots] (0,9pt) -- (0,-\coverdivbot);%
}

\definecolor{summarybg}{HTML}{f5fbff}
\colorlet{codehl}{summarybg}            
\definecolor{summaryborder}{HTML}{C9D2F2}

\newtcolorbox{summarybox}{
    colback=summarybg,
    colframe=summaryborder,
    boxrule=0.5pt,
    arc=2mm,
    outer arc=2mm,
    left=2mm,
    right=2mm,
    top=1.5mm,
    bottom=1.5mm,
    before skip=6pt,
    after skip=6pt
}

\creflabelformat{equation}{#2\textup{#1}#3}

 \newtcolorbox[crefname={note}{notes}]{notebox}[1][]{%
    colback=blue!5!white,
    colframe=blue!10!white,
    sharp corners,
    before upper={{\bfseries Note~\thetcbcounter}.\ },
    #1
}
\newtcolorbox[auto counter, number within=section,crefname={insight}{insights}]{insightbox}[1][]{%
    colback=blue!5!white,
    colframe=blue!10!white,
    sharp corners,
    before upper={{\bfseries Insight~\thetcbcounter}.\ },
    #1
}

\newtcolorbox[auto counter, number within=section,crefname={insight}{insights}]{recommendationbox}[1][]{%
    colback=green!5!white,
    colframe=green!15!white,
    sharp corners,
    before upper={{\bfseries Recommendation~\thetcbcounter}.\ },
    #1
}

\theoremstyle{plain}
\newtheorem{theorem}{Theorem}[section]

\theoremstyle{definition}
\newtheorem{definition}[theorem]{Definition}

\theoremstyle{remark}

\newlength\savewidth
\usepackage{tabulary}
\newcolumntype{x}[1]{>{\centering\arraybackslash}p{#1pt}}

\title{Multi-Token Prediction Done Right}

\author{Jayden Teoh}

\date{MIT}

\begin{document}

\maketitle
{\renewcommand{\thefootnote}{}\footnotetext{Correspondence to: \texttt{jayden\_t@mit.edu}}}

\vfill 
\begin{abstract}
Multi-Token Prediction (MTP) is now a common ingredient in language-model pre-training, as it improves downstream performance and enables faster inference through speculative decoding. We identify a flaw in standard MTP training, in which a residual component of the MTP gradient interferes with both next-token and multi-token learning, increasingly so with scale. Specifically, the MTP cross-entropy gradient can be decomposed exactly into a KL self-distillation term that pulls each MTP head toward the model's own next-token prediction (NTP) distribution and improves speculative decoding, and a residual that has a low signal-to-noise ratio and buries the useful KL signal. We introduce the \textbf{KL gradient--recovery trick}: when two distributions share a linear output head, the gradient of their KL divergence can be recovered exactly by combining their cross-entropy gradients through \emph{straight-through gradient routing}. Unlike minimizing the KL directly, this trick requires no vocabulary-sized logit matrices and is compatible with standard fused cross-entropy kernels. This enables \textbf{Soft Multi-Token Prediction (Soft-MTP)}, a \emph{one-line code change} to MTP that removes the noisy residual and backpropagates only the KL self-distillation signal by reusing the existing NTP and MTP cross-entropy computations. Soft-MTP supports both DeepSeek- and Meta-style MTP, adds no parameters, and incurs no measurable overhead. From 167M to 1.2B parameters, Soft-MTP improves over standard MTP with lower NTP perplexity, faster
speculative decoding and higher zero-shot accuracy at every scale, lower MTP perplexity from 689M onward, and stronger math and code reasoning after fine-tuning at 1.2B. Soft-MTP also matches, if not exceeds, direct KL divergence minimization on most metrics. Soft-MTP is, in effect, a free lunch for MTP optimization. Our code can be found at \url{https://github.com/JaydenTeoh/Soft-MTP}. 
\end{abstract}

\vfill
\begin{figure}[H]
    \centering
    \begin{minipage}[t]{0.334\linewidth}
        \centering
        \coverhead{Soft-MTP illustration}
        \begin{minipage}[c][\coverbodyh][c]{\linewidth}
            \centering
            \begin{tikzpicture}[remember picture]
                \node[inner sep=0pt] (schem) {\includegraphics[width=\linewidth, keepaspectratio]{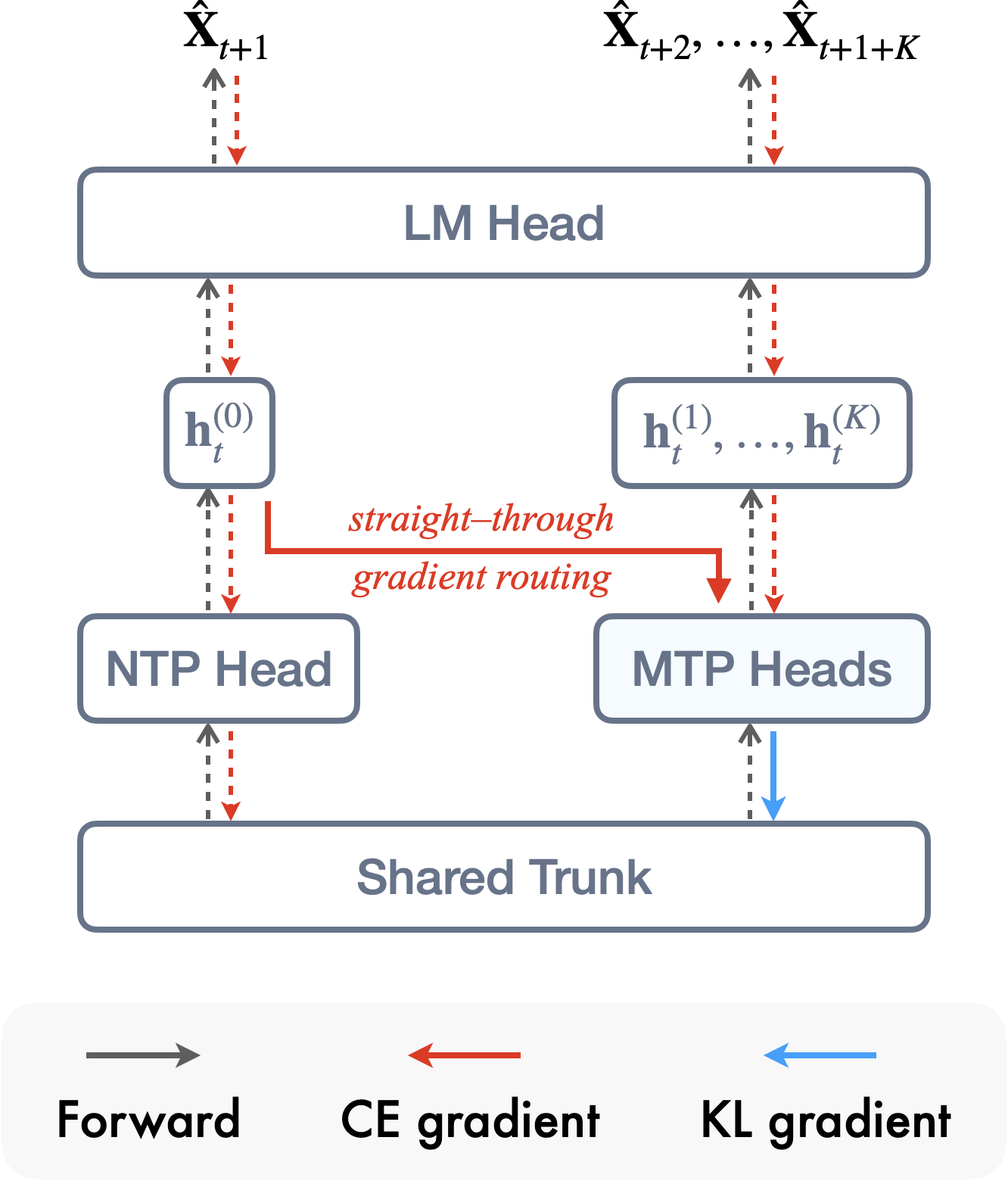}};
                \begin{scope}[shift={(schem.north west)},
                              x={($(schem.north east)-(schem.north west)$)},
                              y={($(schem.south west)-(schem.north west)$)}]
                    \coordinate (lensSrcNW) at (0.255,0.424);   
                    \coordinate (lensSrcSE) at (0.733,0.506);   
                    \coordinate (lensF)     at (0.935,0.465);   
                \end{scope}
                \node[inner sep=0pt, fit=(lensSrcNW)(lensSrcSE)] (lenssrc) {};   
            \end{tikzpicture}
        \end{minipage}
    \end{minipage}%
    \hfill\coverdivider\hfill%
    \begin{minipage}[t]{0.328\linewidth}
        \centering
        \coverhead{One-line code change}
        \begin{minipage}[c][\coverbodyh][c]{\linewidth}
\begin{lstlisting}[style=minimalcode, basicstyle=\ttfamily\codesize\color{codetext}]
z = shared_trunk(x)
h0 = ntp_head(z)

for k in range(1, K+1):
    hk = mtp_head[k-1](z, x)
    loss_mtp += ce(lm_head, hk, y[k])

    (*!\codemark{lensA}!*)# straight-through gradient routing(*!\codemark{lensB}!*)
    # h0 is numerically unchanged
    (*!\hlcode{h0[:, k:] += hk.detach() - hk}\codemark{lensC}!*)

# NTP loss and gradients are unchanged
loss_ntp = ce(lm_head, h0, y[0])

# -ve gradient from NTP loss enters hk
# this recovers KL gradients
(loss_ntp + loss_mtp).backward()
\end{lstlisting}%
\begin{tikzpicture}[remember picture, overlay]
    \node[lensbox, inner sep=1.8pt,
          fit={($(lensA)+(-1.2pt,5.6pt)$) (lensB) ($(lensC)+(0,-4.1pt)$)}] (lenstgt) {};
    \draw[lens] (lenssrc.east) -- (lensF);
    \draw[lens] (lenstgt.north west) -- (lensF) -- (lenstgt.south west);
\end{tikzpicture}
        \end{minipage}
    \end{minipage}%
    \hfill\coverdivider\hfill%
    \begin{minipage}[t]{0.290\linewidth}
        \centering
        \coverhead[0.071\linewidth]{Loss performance}
        \begin{minipage}[c][\coverbodyh][c]{\linewidth}
            \centering
            \includegraphics[width=\linewidth, keepaspectratio]{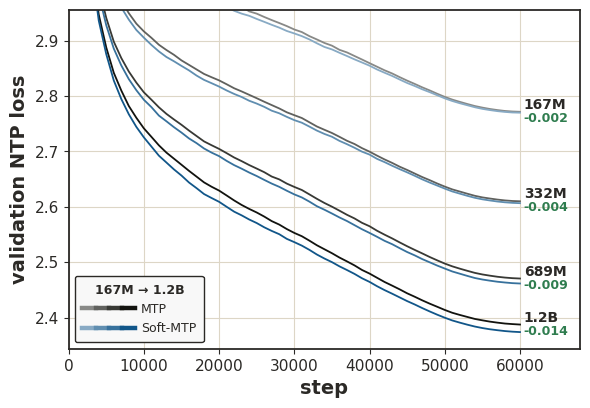}\\[4pt]
            \includegraphics[width=\linewidth, keepaspectratio]{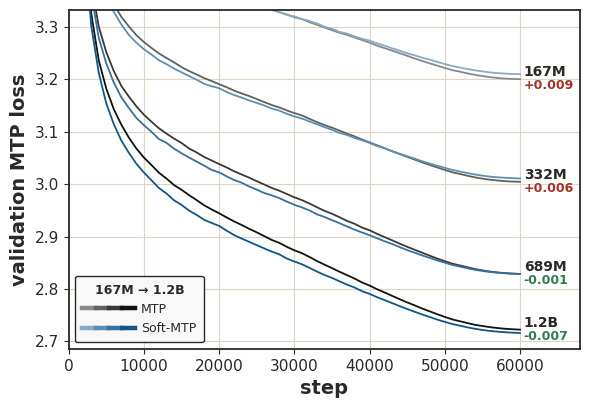}
        \end{minipage}
    \end{minipage}
    \caption{Soft-MTP's illustration, pseudocode, and loss performance compared to MTP. \colorbox{codehl}{\textcolor{codeadd}{Blue line}} is the \textbf{one-line code change}. We refer readers to \Cref{alg:full_soft_mtp_code} for the full code implementation of Soft-MTP, including the MTP loss weighting.} 
    \label{fig:soft_mtp}
\end{figure}
\vfill
\clearpage 


\section{Introduction}

Next-token prediction (NTP) has become a general recipe for training large language models (LLMs), yielding models with remarkable capabilities across a wide range of tasks \citep{radford2019language}. One might argue that the NTP objective is sufficient---any distribution over token sequences can, in theory, be factorized autoregressively and modeled by a sufficiently expressive next-token predictor \citep{sutskever2023nexttoken}. 

Yet, Multi-Token Prediction (MTP) \citep{stern2018blockwise,qi2020prophetnet,gloeckle2024}, which trains a model to predict multiple future tokens, has now been widely adopted in modern LLM training, including in DeepSeek-V3/4 \citep{deepseekai2024deepseekv3technicalreport,deepseekai2026deepseekv4}, Nemotron 3 \citep{nvidia2025nvidianemotron3efficient}, GLM-5 \citep{glm5team2026glm5vibecodingagentic}, and Kimi K3 \citep{kimiteam2026kimik3openfrontier}, just to name a few. MTP offers two main advantages over NTP---better model performance and more efficient inference. As a learning objective, MTP has been shown to alleviate the myopic nature of the next-token supervision \citep{bachmann24a,gloeckle2024,nagarajan2025roll,huang2026how}, leading to stronger downstream performance on long-horizon domains such as mathematics and code. More importantly, MTP heads can also be used for speculative decoding \citep{leviathan2022fast,chen2023acceleratinglargelanguagemodel} to accelerate model inference.

Despite the wide adoption of MTP, we still understand little about how it interacts
with next-token prediction and which objective is best for training the MTP heads. During speculative decoding, a token drafted by MTP is accepted only if it agrees with the model's own next-token prediction. If the goal is to improve speculative decoding, shouldn't we directly train the model's MTP heads to match its own NTP distribution via self-distillation? Recent work does exactly
this \citep{teoh2026nextlatentpredictiontransformerslearn,zhao2026selfdistillationmultitokenprediction},
but naïvely minimizing an explicit KL divergence requires materializing vocabulary-sized logit tensors
for both the teacher and the student, which is costly in memory and incompatible with the standard fused cross-entropy kernels used in modern pre-training \citep{hsu2025ligerkernel,wijmans2025cut}. In practice, frontier models still train their MTP heads with standard cross-entropy against the data.

In this paper, we show that the standard MTP cross-entropy gradient already contains this
self-distillation signal. The MTP gradient decomposes exactly into a KL term $\gsoft$, which
pulls the MTP head toward the model's own next-token distribution, and a residual $\gres$, which
is the NTP cross-entropy gradient against the data (see \cref{sec:soft_mtp}). We provide evidence that this residual in the MTP loss gradient has a low signal-to-noise ratio (SNR) and interferes with both NTP and MTP learning, especially as scale increases. We then
introduce the \emph{KL gradient--recovery trick}: when two distributions share a linear output
head, the gradient of their KL divergence can be recovered exactly from their cross-entropy
gradients through \emph{straight-through gradient routing} \citep{bengio2013estimatingpropagatinggradientsstochastic,vqvae2017}, without explicitly computing the KL or materializing any logits. This trick enables \emph{Soft Multi-Token Prediction (Soft-MTP)}, which removes the residual and sends only the useful KL self-distillation gradient through the MTP heads and the shared model trunk, with a one-line code
change (see \cref{fig:soft_mtp}).

From 167M to 1.2B parameters, Soft-MTP improves over standard MTP with lower NTP perplexity, faster
self-speculative decoding in every domain we test, and higher zero-shot accuracy at every
scale, as well as stronger code and math reasoning after fine-tuning.
It also reaches lower MTP loss from 689M parameters onward. Soft-MTP matches or exceeds the costly explicit KL self-distillation on most metrics, at no
extra cost over standard MTP. Soft-MTP supports both DeepSeek-style \citep{deepseekai2024deepseekv3technicalreport} and Meta-style \citep{gloeckle2024} MTP architectures, adds no parameters, has negligible overhead, and remains compatible with fused cross-entropy kernels and the parallelism used in frontier pre-training. Soft-MTP is,
in effect, a free lunch for MTP optimization.

\section{Related Work}

\paragraph{Multi-token prediction.} Predicting several future tokens dates back to blockwise parallel decoding \citep{stern2018blockwise} and
ProphetNet \citep{qi2020prophetnet}.
\citet{gloeckle2024} showed that MTP with parallel output heads improves large language models, especially on code, and DeepSeek-V3 \citep{deepseekai2024deepseekv3technicalreport} brought MTP to frontier-scale pre-training and introduced the sequential MTP modules now used by many frontier models \citep{deepseekai2024deepseekv3technicalreport,qwen3technicalreport,meituan2025longcatflashtechnicalreport,5team2025glm45agenticreasoningcoding,lingteam2025activationboostedscalinggeneral,nvidia2025nvidianemotron3efficient,coreteam2026mimov2flashtechnicalreport,glm5team2026glm5vibecodingagentic,deepseekai2026deepseekv4,chen2026minimaxm2seriesminiactivations,kimiteam2026kimik3openfrontier}. A line of work argues that MTP helps to counter the myopia of next-token supervision \citep{bachmann24a,gloeckle2024,nagarajan2025roll,huang2026how}. Recent works also go beyond MTP with richer objectives \citep{hu2025the,ahn2025jtp,liu2025lmtp,teoh2026nextlatentpredictiontransformerslearn,Mahajan2026,zuhri2026predicting}.

\paragraph{MTP for speculative decoding.} Speculative decoding drafts several tokens cheaply and verifies them in parallel using the more
expensive target model \citep{leviathan2022fast,chen2023acceleratinglargelanguagemodel}. MTP heads trained during pre-training can serve as drafters for speculative decoding. This is
especially useful for accelerating reinforcement learning post-training, where generating rollouts
is often the main bottleneck \citep{coreteam2026mimov2flashtechnicalreport,li2026breakingentropyboundsaccelerating,iso2026rlacceleration}. Recent work trains drafters to match the verifier more directly, by minimizing the TV
distance \citep{samarin2026lk,li2026breakingentropyboundsaccelerating}, and Nemotron~3
Ultra \citep{nvidia2026nemotron3ultraopen} introduces an \emph{MTP Boosting} post-training phase that
aligns the MTP heads with the NTP distribution via a temperature-scaled forward KL objective. Soft-MTP pursues the same goal during pre-training, at no extra cost over the standard MTP objective.

\paragraph{Self-distillation for MTP.} Recent works train MTP heads to match the NTP distribution. They either compute the KL explicitly \citep{teoh2026nextlatentpredictiontransformerslearn,kirchenbauer2026multitokenpredictionselfdistillation}, which requires materializing the teacher's and the student's distributions over the full vocabulary, or
approximate it with top-$k$ logits \citep{zhao2026selfdistillationmultitokenprediction}. \citet{chang2026klcachetrick} shows a cool trick to reduce the cost of full-vocabulary distillation by caching the teacher's contribution to the KL gradient. However, their method still requires materializing the teacher's full-vocabulary distribution,
and it requires the student's output head to be frozen. We show that standard MTP already contains the self-distillation gradient, and that our KL gradient--recovery trick recovers it exactly from the
existing cross-entropy gradients. Soft-MTP therefore incurs no measurable overhead beyond that of standard MTP and remains compatible with fused cross-entropy
kernels~\citep{hsu2025ligerkernel,wijmans2025cut}.

\section{Background}
Let $x_{1:T}$ denote a sequence of tokens, where each token is represented as a one-hot vector $x_t \in \{0,1\}^V$ over a vocabulary of size $V$. A model trunk with parameters $\theta$ maps each prefix sequence $x_{1:t}$ to a hidden state $\hidden_t^{(0)} \in \mathbb{R}^d$. A linear output head $W_\theta \in \mathbb{R}^{V \times d}$ (also commonly called the \emph{``LM head''}) projects these hidden state representations to logits, followed by a softmax function, to obtain the next-token probability distribution:
\begin{align*}
    p_t^{(0)} = \mathrm{softmax}\left(W_\theta \hidden_{t}^{(0)} \right).
\end{align*}
The standard next-token cross-entropy loss is
\begin{align*}
    \mathcal{L}_\text{NTP} = \sum^{T-1}_{t=1}\mathrm{CE}\left(x_{t+1}, p_t^{(0)}\right)= -\sum^{T-1}_{t=1}\log p_t^{(0)}(x_{t+1}).
\end{align*}

\subsection{Multi-Token Prediction}
Multi-Token Prediction (MTP) extends NTP with $K$ token prediction heads to predict $K$ additional tokens. The $k$-th MTP head produces a hidden state $\hidden_t^{(k)}$ at each position $t$, where $k \in \{1,\dots, K\}$. Following typical implementation, the linear output head $W_\theta$ is \emph{shared} with the MTP heads to generate
\begin{align*}
    p_t^{(k)} = \mathrm{softmax}\left(W_\theta \hidden_{t}^{(k)} \right),
\end{align*}
where $p_t^{(k)}$ predicts the token distribution over $x_{t+k+1}$. The $k$-th MTP loss is then given by
\begin{align*}
    \mathcal{L}_\text{MTP}^{(k)} = \sum^{T-k-1}_{t=1}\mathrm{CE}\left(x_{t+k+1}, p_t^{(k)}\right)= -\sum^{T-k-1}_{t=1}\log p_t^{(k)}(x_{t+k+1}).
\end{align*}
The overall MTP objective is
\begin{align*}
    \mathcal{L}_\text{MTP} = \mathcal{L}_\text{NTP} + \lambda_{\text{MTP}} \cdot \frac{1}{K}\sum_{k=1}^K \mathcal{L}_\text{MTP}^{(k)},
\end{align*}
where $\lambda_{\text{MTP}} > 0$ is a scalar coefficient. For clarity of subsequent derivations, we write each loss as a sum over positions. Implementations
typically average each loss over its own number of targets instead; we account for the resulting
rescaling in \cref{app:loss_weighting} and show the implementation in \cref{alg:full_soft_mtp_code}.

\section{Decomposing the MTP Gradient}
\label{sec:soft_mtp} 

\begin{wrapfigure}{r}{0.35\textwidth}
\centering
\includegraphics[width=\linewidth]{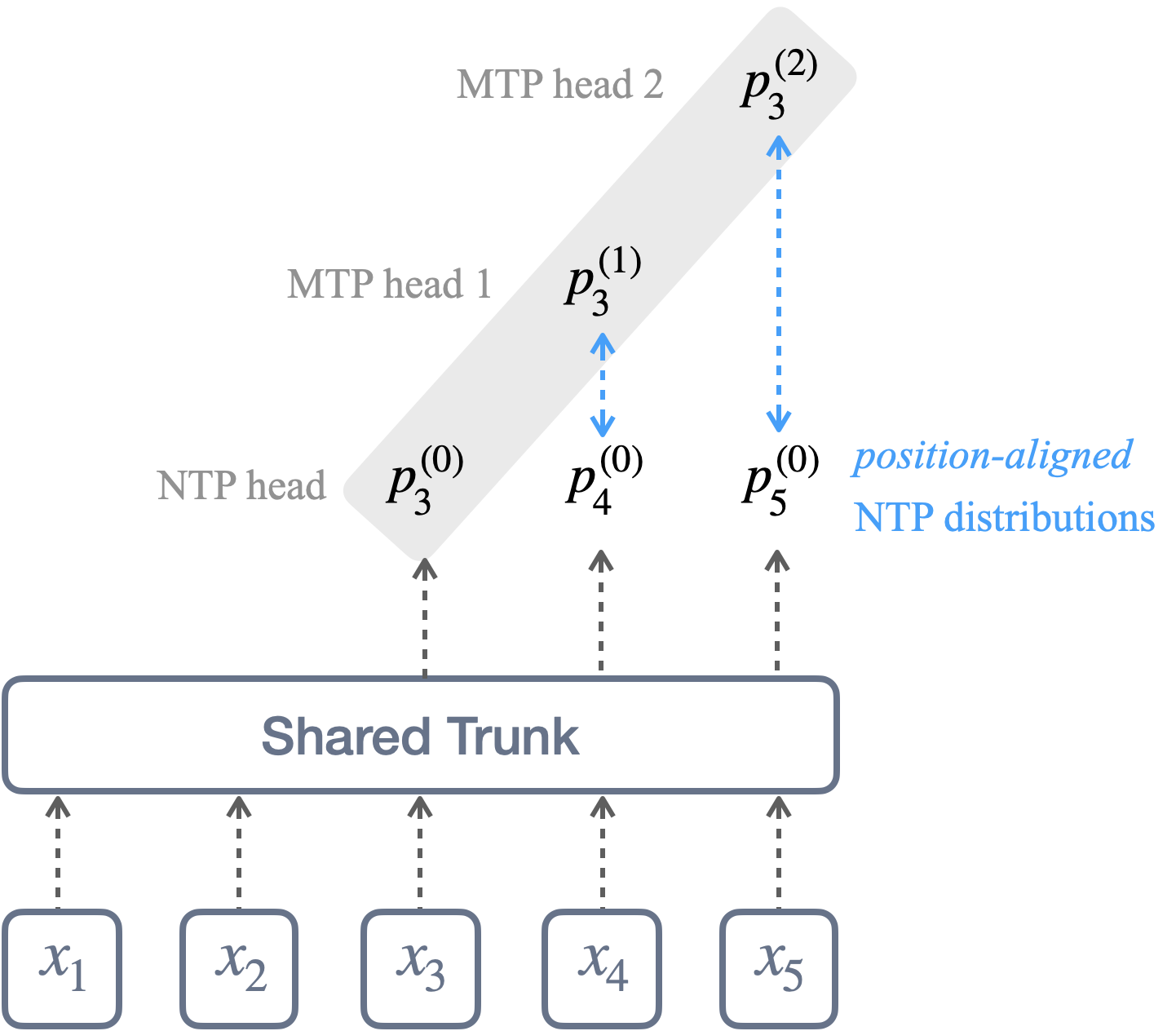}
\caption{Illustration of position-aligned NTP distributions.}
\label{fig:pos_aligned}
\end{wrapfigure}

Note that the analysis that follows is agnostic to the choice of MTP architecture, be it parallel \citep{gloeckle2024} or sequential \citep{deepseekai2024deepseekv3technicalreport}. Consider the $k$-th MTP prediction for position $t$:
\begin{align*}
    p_t^{(k)} = \mathrm{softmax}\left(W_\theta \hidden_{t}^{(k)} \right).
\end{align*}
This distribution predicts $x_{t+k+1}$. The \emph{position-aligned} NTP distribution that predicts the same token $x_{t+k+1}$ is
\begin{align*}
    q_t^{(k)} \coloneqq p_{t+k}^{(0)} = \mathrm{softmax}\left(W_\theta \hidden_{t+k}^{(0)} \right).
\end{align*}
\Cref{fig:pos_aligned} provides a visualization of the position-aligned NTP distributions for an MTP model with $K=2$.
\begin{definition}[Hard MTP gradient] \label{def:hard_mtp_grad}
We define the \emph{hard MTP gradient} as the gradient of the standard MTP cross-entropy loss with respect to the $k$-th MTP hidden state:
\begin{align} 
    g_\text{hard} &\coloneqq \nabla_{\hidden_{t}^{(k)}} \mathrm{CE}\left(x_{t+k+1}, p_t^{(k)}\right) \nonumber \\
    &= W_\theta^\intercal\left(p_t^{(k)} - x_{t+k+1}\right). \label{eq:hard_mtp_grad}
\end{align}
The full derivation can be found in \autoref{app:mtp_grad}. The hard MTP gradient can be further decomposed as such
\begin{align} \label{eq:hard_mtp_grad_decompose}
    g_\text{hard} = W_\theta^\intercal\left(p_t^{(k)} - q_t^{(k)} + q_t^{(k)} -x_{t+k+1}\right)
    = \underbrace{W_\theta^\intercal\left(p_t^{(k)} - q_t^{(k)}\right)}_{\gsoft} + \underbrace{W_\theta^\intercal\left(q_t^{(k)} -x_{t+k+1}\right)}_{\gres},
\end{align}
where $\gsoft$ is a soft \emph{KL self-distillation} component\footnote{We use the term \emph{soft gradients} by analogy to ``soft targets'' in the knowledge distillation literature.} and $\gres$ is a \emph{residual} component corresponding to the cross-entropy gradient with respect to the aligned NTP hidden state. We denote this residual term using $\gres$ to distinguish this MTP gradient decomposition component from $g_\text{NTP}$, which we reserve for the actual NTP cross-entropy gradient propagated through the NTP head into the shared model backbone.
\end{definition}

\subsection{The KL self-distillation component}
Now, we explain why $\gsoft$ corresponds to the KL self-distillation gradient. Treating the aligned NTP distribution $q_t^{(k)}$ as a \emph{stop-gradient} teacher distribution, the forward KL divergence from the aligned NTP distribution to the $k$-th MTP distribution is
\begin{align*}
    D_\mathrm{KL} \left(\stopgrad(q_t^{(k)}) \;\|\;  p_t^{(k)}\right),
\end{align*}
where $\stopgrad(\cdot)$ denotes the stop-gradient operator. The gradient of the KL objective with respect to the MTP hidden state is
\begin{align} \label{eq:kl_grad}
\nabla_{\hidden_{t}^{(k)}}D_\mathrm{KL}\left(\stopgrad(q_t^{(k)}) \| p_t^{(k)}\right) = W_\theta^\intercal \left(p_t^{(k)} - q_t^{(k)}\right) = \gsoft.
\end{align}
We defer the full derivation to \autoref{app:mtp_grad}. Thus, because MTP is supervised by the same underlying data distribution (just with offset token positions) and a shared linear output head $W_\theta$ is used, part of the hard MTP gradient already implicitly performs self-distillation, i.e., it encourages the MTP head to match the aligned NTP distribution. This is also a more sensible target than the one-hot label, because if even the NTP teacher cannot confidently predict $x_{t+k+1}$, there is little reason to expect the MTP head, which typically has far fewer parameters, to predict it $k$ steps earlier. Instead, the NTP teacher's distribution $q_t^{(k)}$ serves as a soft target, as in knowledge distillation \citep{hinton2015distilling}. 

\section{Soft Multi-Token Prediction}
The earlier gradient decomposition leads to the question: how can we retain the self-distillation component $\gsoft$ by removing the residual component $\gres$ from the hard MTP gradient propagated through the shared model trunk?

A naïve implementation would explicitly construct the aligned NTP teacher distribution $q_t^{(k)}$ and $k$-th MTP distribution $p_t^{(k)}$, and minimize $D_\mathrm{KL} \left(\stopgrad(q_t^{(k)}) \;\|\;  p_t^{(k)}\right)$ directly in the logit space. However, given that the vocabulary size $V$ is large in modern language models, storing large $T \times V$ logit matrices during every MTP prediction is costly in memory usage. It is also incompatible with standard fused cross-entropy kernels, which are designed to avoid materializing the entire logit matrix in memory, and are commonly used in language model pre-training. A custom fused KL kernel could avoid materializing the logits, but it would have to recompute the teacher logits that the NTP loss already computes, unless the NTP cross-entropy and all $K$ MTP KL terms were fused into a single custom kernel.

We instead show that the same representation-space KL gradient can be recovered \emph{exactly} from the cross-entropy gradients already computed during standard MTP training, with no measurable overhead. We term this novel procedure of recovering a KL gradient from the difference of two cross-entropy gradients under a shared linear output head as the \emph{KL gradient--recovery trick}. We also show in \cref{subsection:soft_mtp_wins_kl_mtp} that recovering the KL gradient using this trick matches, if not exceeds, direct KL minimization on most metrics.

\subsection{KL gradient--recovery trick}
For clarity, we exclude the $\lambda_\text{MTP}/{K}$ loss weighting from our derivations in this section. Note that our earlier gradient decomposition and the KL gradient--recovery trick are easily extendable to include the loss weighting, which we defer to \cref{app:loss_weighting}. 

From the decomposition derived in \cref{eq:hard_mtp_grad_decompose}, our intuition lies in injecting a $- \gres$ gradient term at the MTP hidden state, so that only $\gsoft$ is propagated through the MTP heads and subsequently the shared model trunk. We start by augmenting the NTP hidden states
\begin{align}\label{eq:straight-through-grad}
    \tilde{\hidden}_{t} = \hidden_{t}^{(0)} + \underbrace{\sum_{\substack{k=1 \\ t-k \geq 1}}^K\left[\stopgrad(\hidden_{t-k}^{(k)}) - \hidden_{t-k}^{(k)}\right]}_{=\;0}.
\end{align}
This resembles the stop-gradient routing used in \emph{straight-through estimators} \citep{bengio2013estimatingpropagatinggradientsstochastic,vqvae2017}, except our intention is not to approximate an unavailable gradient, but rather, analytically derive another gradient quantity. Define the \emph{augmented NTP cross-entropy loss} as
\begin{align} \label{eq:augmented_ntp_loss}
    \tilde{\mathcal{L}}_\text{NTP} = \sum^{T-1}_{t=1}\mathrm{CE}\left(x_{t+1}, \text{softmax}\left(W_\theta \tilde{\hidden}_{t} \right) \right).
\end{align}
Since $\tilde{\hidden}_{t+k}$ is \textbf{numerically the same} as $\hidden_{t+k}^{(0)}$, we have $\text{softmax}\left(W_\theta \tilde{\hidden}_{t+k}\right) = q_t^{(k)}$, and the gradient of the augmented NTP loss with respect to the NTP hidden state remains the same. That is, the model's optimization for the NTP loss remains unaltered. However, it now injects an extra gradient with respect to the MTP representation $\hidden_{t}^{(k)}$
\begin{align}
    \nabla_{\hidden_{t}^{(k)}}\tilde{\mathcal{L}}_\text{NTP} &= -\nabla_{\tilde{\hidden}_{t+k}}\mathrm{CE}\left(x_{t+k+1}, \text{softmax}\left(W_\theta \tilde{\hidden}_{t+k} \right)\right) \nonumber \\
    &= -W_\theta^\intercal\left(q_t^{(k)} -x_{t+k+1}\right) \nonumber \\
    &= -\gres, \label{eq:injected_grad}
\end{align}
which is precisely the quantity we needed. Therefore, summing the contributions of the hard MTP gradient from the ordinary MTP loss and the gradient of the augmented NTP loss with respect to the MTP hidden states, we get the self-distillation gradients
\begin{align} \label{eq:recovered_kl_grad}
    \boxed{\nabla_{\hidden_{t}^{(k)}}\left(\tilde{\mathcal{L}}_\text{NTP} + \mathcal{L}^{(k)}_\text{MTP} \right) = g_\text{hard} - \gres= \gsoft}
\end{align}
as derived in our decomposition in \cref{eq:hard_mtp_grad_decompose}. \Cref{fig:soft_mtp} illustrates the straight-through gradient routing and the \textbf{one-line code change} needed to operationalize the KL gradient--recovery. In our implementation, we also verify that the recovered gradient matches the explicit KL gradient at every MTP head to machine precision in \texttt{float64}.

\subsection{Soft-MTP}
Our method, \emph{Soft Multi-Token Prediction (Soft-MTP)}, uses the KL gradient--recovery trick to optimize the parameters of the MTP heads and the shared model trunk. There are two main distinctions of Soft-MTP compared to prior self-distillation MTP approaches. 

First, the KL gradient--recovery trick operates directly in the NTP hidden-state space (see \cref{eq:straight-through-grad}), and avoids the need to materialize any additional logit matrices or explicitly compute KL divergence. As a result, Soft-MTP introduces negligible computational and memory overhead over standard MTP.

Second, Soft-MTP still sends cross-entropy gradients from the MTP loss into the shared linear output head $W_\theta$, but the parameters of the upstream modules (i.e., MTP heads and shared model trunk) receive the self-distillation gradient $\gsoft$ instead. We explain why this is better than direct KL minimization in \cref{subsection:soft_mtp_wins_kl_mtp}. \cref{tab:gradient_routing} presents the anatomy of the gradient routing across different MTP objectives.
\begin{table}[H]
    \centering
    \begin{tabular}{lcc}
        \toprule
        Method
        & Output head $W_\theta$
        & MTP heads and shared trunk \\
        \midrule
        Standard MTP
        & CE gradient
        & CE gradient \\
        KL-MTP
        & KL gradient
        & KL gradient \\
        \textbf{Soft-MTP}
        & CE gradient
        & KL gradient \\
        \bottomrule
    \end{tabular}
    \caption{Gradient routing under different MTP objectives. We omit the standard NTP gradient which is unchanged across methods.}
    \label{tab:gradient_routing}
\end{table}

\begin{summarybox}
\textbf{Note:} Please refer to \autoref{app:full_pseudocode} for the full
PyTorch-style Soft-MTP code implementation using DeepSeek MTP, including the $\lambda_\text{MTP}/K$ loss weighting and the rescaling that accounts for different numbers of targets across MTP heads.
\end{summarybox}

\subsection{A bias--variance view of the residual}
\label{subsec:bias_variance}

\Cref{eq:hard_mtp_grad_decompose} can also be interpreted from a bias--variance decomposition perspective. Of its two components, only $\gres$ depends on the sampled target token $x_{t+k+1}$; $\gsoft$ is determined entirely by the model's own NTP and MTP predictions. Let $\pi_t^{(k)}$ denote the true distribution of the next token $x_{t+k+1}$ given the prefix $x_{\le t+k}$, which is exactly what the position-aligned NTP distribution $q_t^{(k)}$ tries to predict. Since the one-hot target is a sample $x_{t+k+1} \sim \pi_t^{(k)}$, the residual splits into two parts:
\begin{align}\label{eq:bv_residual_split}
    \gres = W_\theta^\intercal\left(q_t^{(k)} - x_{t+k+1}\right)
    = \underbrace{W_\theta^\intercal\left(q_t^{(k)} - \pi_t^{(k)}\right)}_{\text{NTP error}}
    + \underbrace{W_\theta^\intercal\left(\pi_t^{(k)} - x_{t+k+1}\right)}_{\text{one-hot sampling noise}},
\end{align}
where the sampling noise has zero mean because $\mathbb{E}[x_{t+k+1}] = \pi_t^{(k)}$. The hard MTP gradient is unbiased but it includes the $\gres$ component and incurs the full one-hot sampling variance. By removing $\gres$ from the MTP updates, Soft-MTP removes this variance at the cost of a bias: without the
NTP-error correction term, $\gsoft$ is biased toward the NTP distribution and not the data distribution $\pi_t^{(k)}$.
These two terms behave very differently as models scale. The bias term is the NTP teacher's own prediction error, which keeps \emph{shrinking} as the base model scales and/or trains longer. The variance term is the \emph{irreducible} noise of sampling a one-hot target, which no amount of scale or training removes. Therefore, the trade that Soft-MTP makes becomes increasingly favorable at scale. We observe this empirically in \cref{subsection:effect_of_residual}, where the residual has a much lower \emph{signal-to-noise} ratio than the KL component, and the gap widens with model scale. Furthermore, since MTP is largely used to accelerate inference, biasing the MTP heads towards matching the NTP distribution is actually desirable for improving speculative decoding acceptance\footnote{Technically, the acceptance rate is determined by the TV distance and KL provides an indirect upper bound through Pinsker's inequality (see \citet{samarin2026lk} and \citet{li2026breakingentropyboundsaccelerating}). Nonetheless, KL is the more appropriate training loss here: it is smooth where TV is not, and it keeps a non-vanishing gradient on tokens the MTP heads underpredict.}. Unlike KL-MTP, Soft-MTP still trains the shared output head $W_\theta$ with cross-entropy, so the
head is fitted to the data rather than biased toward its own next-token predictions. We explain
why this matters in \cref{subsection:soft_mtp_wins_kl_mtp}.

\section{Experiments}

We now turn to an empirical validation of Soft-MTP. Our study is organized around three main questions:

\begin{enumerate}
    \item \textbf{How do the self-distillation component $\gsoft$ and residual component $\gres$ of the MTP cross-entropy gradient affect next-token and multi-token learning?}
    
    In \cref{subsection:effect_of_residual}, we find that $\gres$ has a substantially lower \emph{signal-to-noise ratio (SNR)} than $\gsoft$, yet accounts for a large fraction of the MTP gradient norm. Moreover, this residual becomes increasingly noisy with model scale, obscuring the useful learning signal in the MTP cross-entropy gradient. We further find that the residual component is anti-aligned with the NTP update direction throughout training, suggesting that the residual not only adds noise to MTP learning but also interferes with NTP optimization. Soft-MTP reaches a lower NTP loss than standard MTP, and the gap widens with scale. From 689M parameters onward, it also reaches a lower MTP loss.

    \item \textbf{Does isolating the self-distillation gradient in Soft-MTP improve over standard MTP?}
    
    In \cref{subsection:soft_mtp_wins_hard_mtp}, we empirically validate the analysis above by comparing Soft-MTP against standard MTP. Across model scales, Soft-MTP yields better perplexity, downstream accuracy, speculative-decoding acceptance rates, and math and code fine-tuning performance. These results support our analyses in Question 1---removing the noisy residual produces a cleaner and more compatible learning signal.

    \item \textbf{How does Soft-MTP compare with directly optimizing a KL self-distillation objective?}
    
    In \cref{subsection:soft_mtp_wins_kl_mtp}, we find little benefit from explicitly computing the KL divergence or propagating its gradients through the LM head: retaining hard cross-entropy supervision at the LM head while propagating the recovered KL gradients only through the MTP heads and shared model backbone is sufficient to match or exceed the gains of direct self-distillation. Moreover, Soft-MTP avoids the computational overhead of computing KL divergence and preserves compatibility with fused linear cross-entropy kernels. Overall, this suggests that Soft-MTP is more efficient.
\end{enumerate}

\paragraph{Experimental setup.} We pre-trained LLaMA-style models \citep{touvron2023llama} of base-model sizes ranging from 167M to 1.2B parameters using NTP and MTP objectives. For MTP, we used the widely adopted DeepSeek-style sequential architecture \citep{deepseekai2024deepseekv3technicalreport} and added three MTP heads to predict $K=3$ additional future tokens, following common practice \citep{gloeckle2024,coreteam2026mimov2flashtechnicalreport, chen2026minimaxm2seriesminiactivations, glm5team2026glm5vibecodingagentic}. To study the effect of standard MTP and Soft-MTP consistently across model sizes, we kept each MTP head at a constant 3.2\% of the trunk's non-embedding parameters and fixed the MTP loss coefficient at $\lambda_{\text{MTP}}=1.0$. We trained all models on the FineWeb-Edu dataset \citep{penedo2024fineweb} for 60,000 steps with a batch size of 0.5M tokens and a sequence length of 2048, for a total of approximately 31.5B training tokens, which is $\sim1.3\times$ Chinchilla-optimal \citep{hoffmann2022training} for our largest model size. To be of modern relevance, we trained our models using the Muon optimizer \citep{jordan2024muon} under the update-RMS rescaling of \citet{liu2025muon}. Our training hyperparameters are detailed in \autoref{app:exp_hyperparam}. For completeness, we also showed that Soft-MTP improves over standard MTP using Meta-style MTP heads in \autoref{app:meta_mtp_results}.

\subsection{The residual interferes with learning} \label{subsection:effect_of_residual}

\begin{figure}[htbp]
     \centering
     \begin{subfigure}[b]{0.33\textwidth}
         \centering
         \includegraphics[width=\textwidth, keepaspectratio]{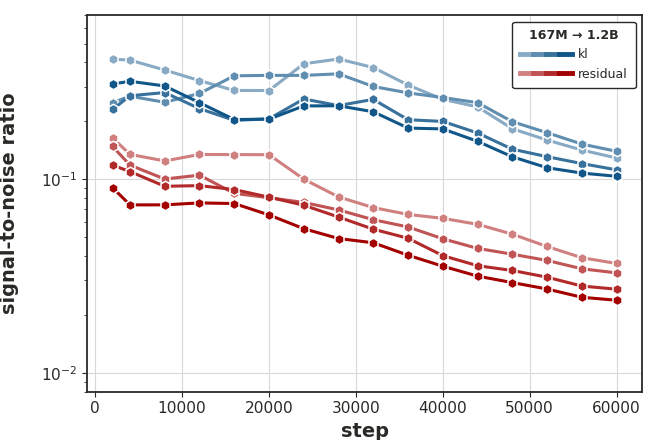}
         \caption{\textbf{167M--1.2B:} Trunk + MTP gradient SNR.}
         \label{fig:all_scales_trunk_mtp_head_snr}
     \end{subfigure}
     \hfill
     \begin{subfigure}[b]{0.33\textwidth}
         \centering
         \includegraphics[width=\textwidth, keepaspectratio]{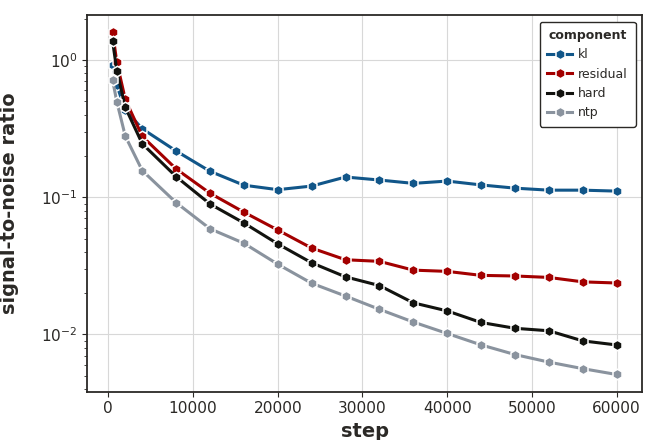}
         \caption{\textbf{1.2B:} Trunk gradient SNR.}
         \label{fig:1p2b_trunk_snr}
     \end{subfigure}
     \hfill
     \begin{subfigure}[b]{0.33\textwidth}
         \centering
         \includegraphics[width=\textwidth, keepaspectratio]{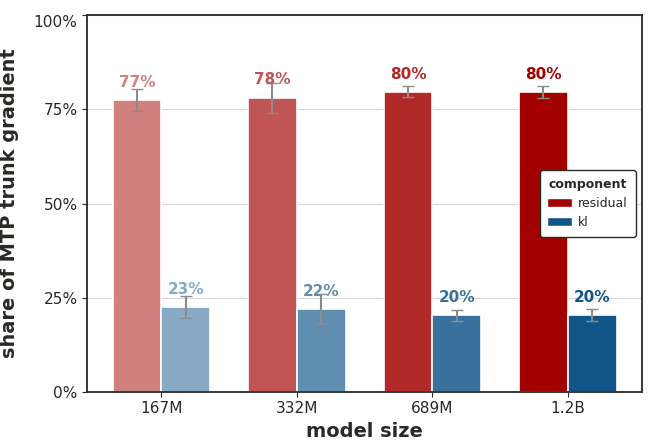}
         \caption{\textbf{167M--1.2B:} Gradient share in trunk.}
         \label{fig:mtp_grad_share}
     \end{subfigure}
     \caption{\textbf{The residual has substantially lower SNR than the KL gradient but dominates the MTP gradient signal.} (a-b) Signal-to-noise ratio (SNR) of each gradient component with respect to the shared trunk and MTP head parameters, across scales. (c) Share of the MTP gradient in the shared trunk carried by each component, averaged over training steps; the residual carries $\sim$80\% at every scale.}
     \label{fig:gradients_snr}
\end{figure} 

\paragraph{Gradient signal-to-noise.} We first measure the signal-to-noise ratios (SNR) of the different gradient components w.r.t. the model parameters $\theta$: $g^\theta_\text{hard}$ (hard MTP gradient), $\gsoft[\theta]$ (KL self-distillation gradient), $\gres[\theta]$ (residual MTP gradient), and $g^\theta_\text{NTP}$ (NTP cross-entropy gradient). For a gradient component $g^\theta_c(\xi)$ computed on a data batch $\xi$, we define its \emph{signal} as the squared norm of its expected direction, $S_c = \lVert \mathbb{E}_\xi[g^\theta_c] \rVert^2$, and its \emph{noise} as the total variance around this direction, $N_c = \mathbb{E}_\xi\lVert g^\theta_c \rVert^2 - S_c = \operatorname{tr}(\Sigma_c)$. We report $\mathrm{SNR}_c = S_c / N_c$ as the signal-to-noise ratio. This quantity is closely related to the inverse of the \emph{gradient noise scale} of \citet{mccandlish2018}; with a fixed batch size, the two differ only by a constant factor. We estimate the SNR using $M=256$ independent batches, each containing 4 sequences of length 2048. The naïve signal estimate is biased upward by $\operatorname{tr}(\Sigma)/M$, so we report the unbiased estimator following \citet[Appendix A.1]{mccandlish2018}. 


\Cref{fig:all_scales_trunk_mtp_head_snr,fig:1p2b_trunk_snr} compare the SNR of the two gradient components w.r.t. the shared trunk and MTP head parameters across model scales. Throughout training, \textbf{the residual carries roughly $\mathbf{4\times}$ lower SNR than the KL component}. The residual becomes increasingly noisy, and the gap widens as the scale increases, from $4.1\times$ at 167M to $4.6\times$ at 1.2B (note the log-scale axis). 
This result corroborates the bias--variance tradeoff discussed in \cref{subsec:bias_variance}: the residual always contains irreducible noise from sampling the one-hot target, while its signal, i.e., the NTP error, shrinks with scale, so its SNR falls as models grow. Removing the residual substantially increases gradient SNR---the KL component has much higher SNR than the original hard-MTP gradient. Yet despite its low SNR, the residual accounts for $\sim$80\% of the MTP gradient reaching the shared trunk (see \cref{fig:mtp_grad_share}). Thus, although the standard MTP contains the high SNR KL signal implicitly, the residual obscures it under high-variance updates, which is why
the overall hard-MTP gradient has lower SNR (see \cref{fig:1p2b_trunk_snr}). We detail how we measure the SNR and gradient share of the KL and residual components in \autoref{app:diagnostics}.

\begin{figure}[htbp]
     \centering
     \begin{subfigure}[b]{0.33\textwidth}
         \centering
         \includegraphics[width=\textwidth, keepaspectratio]{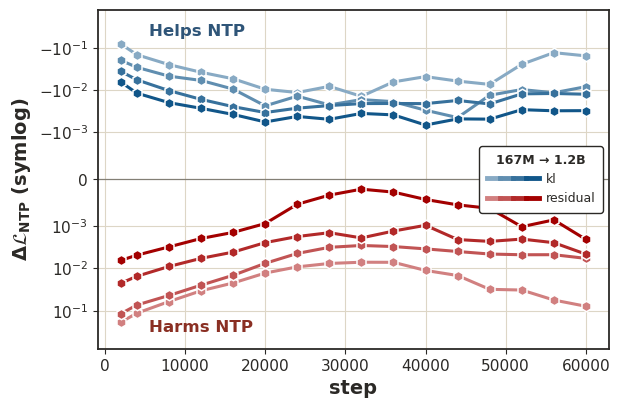}
         \caption{\textbf{167M--1.2B:} Effect on NTP loss (proxy).}
         \label{fig:ntp_help_harm}
     \end{subfigure}
     \hfill
     \begin{subfigure}[b]{0.33\textwidth}
         \centering
         \includegraphics[width=\textwidth, keepaspectratio]{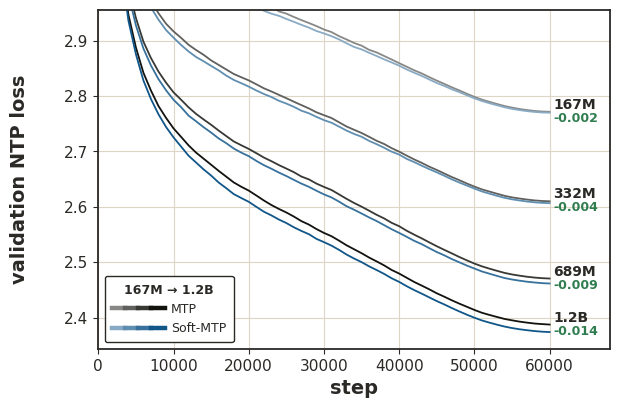}
         \caption{\textbf{167M--1.2B:} NTP loss.}
         \label{fig:ntp_loss_curve}
     \end{subfigure}
     \hfill
     \begin{subfigure}[b]{0.33\textwidth}
         \centering
         \includegraphics[width=\textwidth, keepaspectratio]{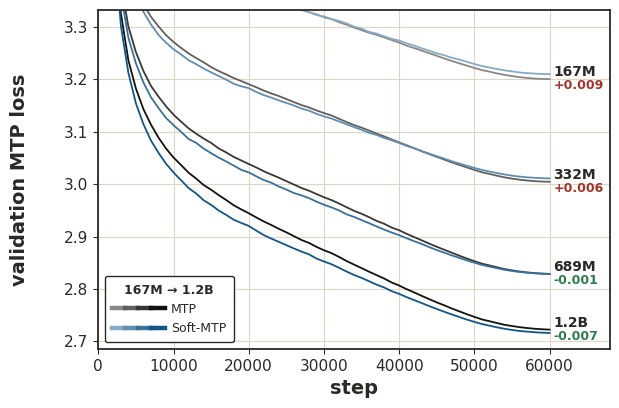}
         \caption{\textbf{167M--1.2B:} MTP loss.}
         \label{fig:mtp_loss_curve}
     \end{subfigure}
     \caption{\textbf{The residual worsens NTP and MTP loss especially as scale increases.} (a) Change in NTP loss $\Delta\mathcal{L}_{\text{NTP}}$ from a unit step of the trunk hidden state representation along each gradient component across model scales. (b-c) NTP and MTP loss of standard MTP and Soft-MTP throughout training across scales.}
     \label{fig:ntp_harm_figs}
\end{figure} 

\paragraph{Interaction with NTP and MTP losses.} Earlier, we showed the residual has low SNR. Here, we aim to show that the residual is not only noisy, but also interferes with NTP and MTP learning. In \cref{fig:ntp_help_harm}, we take a unit step on the shared trunk's hidden state along each component's gradient (backpropagated from
the MTP heads) and re-evaluate the NTP loss through the frozen output head. This is a proxy for what each component `asks' the trunk output to become. A step along the KL component consistently decreases NTP loss, indicating that the self-distillation signal in MTP reinforces the NTP objective. In contrast, a step along the residual component in MTP consistently increases NTP loss. This behavior persists throughout training and is consistent across model scales. Note that this analysis serves only as a characterization of the components' behavior at the trunk--MTP head interface; the realized update to the trunk's hidden states during training also depends on the optimizer used. We empirically corroborate this residual interference in \cref{fig:ntp_loss_curve}: \textbf{removing the residual with Soft-MTP lowers the NTP loss, especially as
scale increases}. Furthermore, from 689M parameters onward, \textbf{Soft-MTP also reaches a lower MTP loss than standard MTP} (see \cref{fig:mtp_loss_curve}).

\begin{summarybox}
\textbf{Summary:} The residual is much noisier (lower SNR) than the KL component, yet accounts for roughly 80\% of the MTP gradient. Removing the residual gradient improves NTP and MTP learning, especially as scale increases.
\end{summarybox}

\subsection{Soft-MTP outperforms standard MTP} \label{subsection:soft_mtp_wins_hard_mtp}

\begin{table}[htbp]
\centering
\definecolor{gaingreen}{HTML}{2E7D32}
\newcommand{\gain}[1]{\,{\scriptsize\textcolor{gaingreen}{+#1\%}}}
\resizebox{\textwidth}{!}{%
\begin{tabular}{cl|cc|cc|cc|cc|cc|cc}
\toprule
\multirow{2}{*}{\rotatebox[origin=c]{90}{\small \textbf{Size}}} & \multirow{2}{*}{\textbf{Model}}
 & \multicolumn{2}{c|}{\textbf{Verifier match}}
 & \multicolumn{2}{c}{\textbf{FineWeb-Edu}}
 & \multicolumn{2}{c}{\textbf{Wikipedia}}
 & \multicolumn{2}{c}{\textbf{Books}}
 & \multicolumn{2}{c}{\textbf{Science}}
 & \multicolumn{2}{c}{\textbf{Math}} \\
\cmidrule(lr){3-4} \cmidrule(lr){5-6} \cmidrule(lr){7-8} \cmidrule(lr){9-10} \cmidrule(lr){11-12} \cmidrule(lr){13-14}
 &
& KL $\downarrow$ & TV $\downarrow$
& Speedup & Acc.\ Len.
& Speedup & Acc.\ Len.
& Speedup & Acc.\ Len.
& Speedup & Acc.\ Len.
& Speedup & Acc.\ Len. \\
\midrule
\multirow{2}{*}{167M} & MTP & 0.485 & 0.305 & 1.90\hphantom{\gain{1.1}} & 1.52\hphantom{\gain{1.3}} & 1.88\hphantom{\gain{0.5}} & 1.49\hphantom{\gain{0.7}} & 1.88\hphantom{\gain{1.6}} & 1.49\hphantom{\gain{2.0}} & 1.88\hphantom{\gain{1.1}} & 1.49\hphantom{\gain{1.3}} & 1.76\hphantom{\gain{2.8}} & 1.36\hphantom{\gain{3.7}} \\
 & Soft-MTP & \textbf{0.470} & \textbf{0.300} & \textbf{1.92}\gain{1.1} & \textbf{1.54}\gain{1.3} & \textbf{1.89}\gain{0.5} & \textbf{1.50}\gain{0.7} & \textbf{1.91}\gain{1.6} & \textbf{1.52}\gain{2.0} & \textbf{1.90}\gain{1.1} & \textbf{1.51}\gain{1.3} & \textbf{1.81}\gain{2.8} & \textbf{1.41}\gain{3.7} \\
\midrule
\multirow{2}{*}{332M} & MTP & 0.448 & 0.288 & 1.96\hphantom{\gain{1.0}} & 1.57\hphantom{\gain{1.9}} & 1.94\hphantom{\gain{0.5}} & 1.54\hphantom{\gain{0.6}} & 1.94\hphantom{\gain{1.5}} & 1.55\hphantom{\gain{1.9}} & 1.94\hphantom{\gain{1.5}} & 1.54\hphantom{\gain{2.6}} & 1.87\hphantom{\gain{3.2}} & 1.47\hphantom{\gain{4.1}} \\
 & Soft-MTP & \textbf{0.430} & \textbf{0.282} & \textbf{1.98}\gain{1.0} & \textbf{1.60}\gain{1.9} & \textbf{1.95}\gain{0.5} & \textbf{1.55}\gain{0.6} & \textbf{1.97}\gain{1.5} & \textbf{1.58}\gain{1.9} & \textbf{1.97}\gain{1.5} & \textbf{1.58}\gain{2.6} & \textbf{1.93}\gain{3.2} & \textbf{1.53}\gain{4.1} \\
\midrule
\multirow{2}{*}{689M} & MTP & 0.410 & 0.271 & 1.88\hphantom{\gain{1.1}} & 1.62\hphantom{\gain{2.5}} & 1.84\hphantom{\gain{1.6}} & 1.58\hphantom{\gain{1.9}} & 1.85\hphantom{\gain{1.6}} & 1.60\hphantom{\gain{2.5}} & 1.86\hphantom{\gain{1.1}} & 1.60\hphantom{\gain{1.9}} & 1.81\hphantom{\gain{1.7}} & 1.57\hphantom{\gain{1.9}} \\
 & Soft-MTP & \textbf{0.390} & \textbf{0.265} & \textbf{1.90}\gain{1.1} & \textbf{1.66}\gain{2.5} & \textbf{1.87}\gain{1.6} & \textbf{1.61}\gain{1.9} & \textbf{1.88}\gain{1.6} & \textbf{1.64}\gain{2.5} & \textbf{1.88}\gain{1.1} & \textbf{1.63}\gain{1.9} & \textbf{1.84}\gain{1.7} & \textbf{1.60}\gain{1.9} \\
\midrule
\multirow{2}{*}{1.2B} & MTP & 0.384 & 0.260 & 1.74\hphantom{\gain{1.1}} & 1.67\hphantom{\gain{1.8}} & 1.72\hphantom{\gain{1.2}} & 1.63\hphantom{\gain{1.8}} & 1.71\hphantom{\gain{1.8}} & 1.62\hphantom{\gain{3.1}} & 1.72\hphantom{\gain{1.7}} & 1.65\hphantom{\gain{1.8}} & 1.70\hphantom{\gain{1.8}} & 1.60\hphantom{\gain{2.5}} \\
 & Soft-MTP & \textbf{0.364} & \textbf{0.253} & \textbf{1.76}\gain{1.1} & \textbf{1.70}\gain{1.8} & \textbf{1.74}\gain{1.2} & \textbf{1.66}\gain{1.8} & \textbf{1.74}\gain{1.8} & \textbf{1.67}\gain{3.1} & \textbf{1.75}\gain{1.7} & \textbf{1.68}\gain{1.8} & \textbf{1.73}\gain{1.8} & \textbf{1.64}\gain{2.5} \\
\bottomrule
\end{tabular}%
}
\caption{Self-speculative decoding performance across domains. \textbf{KL divergence / TV distance:} teacher-forced
divergence of the MTP distribution from the model's own next-token verifier on FineWeb-Edu validation set
(mean over the three MTP heads). \textbf{Speedup:} improvement in inference speed using self-speculative decoding (relative to autoregressive decoding). \textbf{Acceptance Length (Acc.\ Len.):} average accepted tokens per draft, excluding the next-token prediction which is always accepted. \textcolor{gaingreen}{Green:} relative improvement of Soft-MTP over MTP. We \textbf{bold} the better of MTP and Soft-MTP per column; Soft-MTP wins every
metric at every scale.}
\label{tab:spec_eval_results}
\end{table}

\paragraph{Speculative decoding.} We evaluated the speculative decoding performance of the standard MTP and Soft-MTP pre-trained models across FineWeb-Edu, Wikipedia, Books, Science, and Math. For each domain, we sample 256 prompts of length 512 tokens and generate 512-token continuations using the speculative sampling algorithm of \citet{leviathan2022fast}, drafting with the $K=3$ MTP heads and verifying using the model's own next-token head. For each model, we report the acceptance length (average number of accepted tokens per verification step) and the inference speedup relative to standard autoregressive sampling from the base model. \Cref{tab:spec_eval_results} shows that \textbf{Soft-MTP produces better drafts and faster decoding than standard MTP at every scale and in every domain.} Averaged over the five
domains, it improves acceptance length by 1.8--2.2\% and speedup by 1.4--1.5\% at each scale. Its MTP distributions are closer to the model's own NTP verifier, with lower KL divergence and TV distance at every scale. These show that the bias discussed in
\cref{subsec:bias_variance} works in our favor: the self-distillation gradient pulls the MTP heads toward exactly the NTP distribution that verifies their drafts. 

\begin{table}[H]
\centering

\resizebox{\columnwidth}{!}{%
\begin{tabular}{cl|ccc|ccccccccc}
\toprule

\rotatebox[origin=c]{90}{\small \textbf{Size}}
& \textbf{Model}
& \begin{tabular}{@{}c@{}}\textbf{FW-Edu} \\ ppl $\downarrow$\end{tabular}
& \begin{tabular}{@{}c@{}}\textbf{Wiki.} \\ ppl $\downarrow$\end{tabular}
& \begin{tabular}{@{}c@{}}\textbf{LAMB.} \\ ppl $\downarrow$\end{tabular}
& \begin{tabular}{@{}c@{}}\textbf{LAMB.} \\ acc $\uparrow$\end{tabular}
& \begin{tabular}{@{}c@{}}\textbf{PIQA} \\ acc $\uparrow$\end{tabular}
& \begin{tabular}{@{}c@{}}\textbf{HellaS.} \\ acc\_n $\uparrow$\end{tabular}
& \begin{tabular}{@{}c@{}}\textbf{Wino.} \\ acc $\uparrow$\end{tabular}
& \begin{tabular}{@{}c@{}}\textbf{ARC-e} \\ acc $\uparrow$\end{tabular}
& \begin{tabular}{@{}c@{}}\textbf{ARC-c} \\ acc\_n $\uparrow$\end{tabular}
& \begin{tabular}{@{}c@{}}\textbf{OBQA} \\ acc $\uparrow$\end{tabular}
& \begin{tabular}{@{}c@{}}\textbf{SciQ} \\ acc\_n $\uparrow$\end{tabular}
& \textbf{Avg.} \\
\midrule

\multirow{3}{*}{\rotatebox[origin=c]{90}{\small 167M}}
& NTP
& 15.230 & 27.311 & 42.157
& 34.66
& 66.87
& 40.63
& 52.01
& 58.46
& 28.33
& 23.80
& 74.60
& 47.42 \\

& MTP
& 15.983 & 29.074 & \textbf{43.013}
& 33.88
& \textbf{66.43}
& \textbf{39.70}
& 51.70
& \textbf{59.01}
& 26.37
& \textbf{23.40}
& \textbf{74.20}
& 46.84 \\

& \textbf{Soft-MTP}
& \textbf{15.952} & \textbf{28.318} & 43.790
& \textbf{34.12}
& 66.10
& 39.55
& \textbf{53.20}
& 58.84
& \textbf{27.65}
& 22.20
& 73.60
& \textbf{46.91} \\

\midrule

\multirow{3}{*}{\rotatebox[origin=c]{90}{\small 332M}}
& NTP
& 13.073 & 22.482 & 26.005
& 38.54
& 69.91
& 46.65
& 53.51
& 63.51
& 31.40
& 24.80
& 78.00
& 50.79 \\

& MTP
& 13.602 & 23.856 & \textbf{26.997}
& \textbf{38.77}
& \textbf{69.59}
& \textbf{45.92}
& \textbf{54.06}
& 63.01
& 31.23
& \textbf{22.20}
& 78.30
& 50.38 \\

& \textbf{Soft-MTP}
& \textbf{13.551} & \textbf{23.249} & 28.566
& 37.45
& 68.72
& 45.58
& 53.59
& \textbf{63.89}
& \textbf{32.17}
& \textbf{22.20}
& \textbf{80.10}
& \textbf{50.46} \\

\midrule

\multirow{3}{*}{\rotatebox[origin=c]{90}{\small 689M}}
& NTP
& 11.442 & 19.157 & 18.079
& 43.51
& 71.55
& 52.61
& 56.12
& 69.19
& 34.98
& 27.00
& 83.40
& 54.79 \\

& MTP
& 11.834 & 20.265 & 18.560
& \textbf{43.31}
& \textbf{71.11}
& \textbf{52.33}
& 56.59
& \textbf{69.02}
& 35.67
& 25.20
& 82.50
& 54.47 \\

& \textbf{Soft-MTP}
& \textbf{11.724} & \textbf{19.316} & \textbf{17.977}
& 43.30
& 70.62
& 52.28
& \textbf{56.67}
& 68.77
& \textbf{36.35}
& \textbf{26.20}
& \textbf{85.90}
& \textbf{55.01} \\

\midrule

\multirow{3}{*}{\rotatebox[origin=c]{90}{\small 1.2B}}
& NTP
& 10.572 & 17.503 & 14.159
& 46.81
& 73.12
& 55.95
& 57.22
& 71.17
& 37.46
& 27.20
& 85.10
& 56.75 \\

& MTP
& 10.891 & 18.156 & 14.996
& 45.86
& 71.55
& \textbf{55.91}
& 57.46
& 70.75
& \textbf{37.97}
& 29.60
& 85.30
& 56.80 \\

& \textbf{Soft-MTP}
& \textbf{10.741} & \textbf{17.489} & \textbf{14.073}
& \textbf{46.48}
& \textbf{72.03}
& \textbf{55.91}
& \textbf{58.01}
& \textbf{71.97}
& 37.54
& \textbf{30.60}
& \textbf{85.90}
& \textbf{57.31} \\
\bottomrule
\end{tabular}%
}

\caption{Downstream language modeling evaluation. We \textbf{bold} the stronger result between standard MTP and Soft-MTP for every model scale (NTP results are only for reference).}
\label{tab:lm_eval_results}
\end{table}

\paragraph{Zero-shot downstream performance.} \Cref{tab:lm_eval_results}
reports the perplexity (ppl) and zero-shot accuracy (acc, or length-normalized acc\_n) across benchmarks measured using the LM Evaluation Harness~\citep{eval-harness}. For reference, we also include the results of an NTP baseline. Consistent with \citet{gloeckle2024}, standard MTP does not improve multiple-choice accuracy over NTP at small scales. \textbf{Soft-MTP, in contrast, improves average accuracy over standard MTP at every scale}, and the gain becomes larger at scale: around $0.1$ points at 167M and 332M, and around $0.5$ points at 689M and 1.2B, where \textbf{Soft-MTP also surpasses NTP}. At 1.2B, Soft-MTP achieves the best WikiText and LAMBADA perplexity and the highest average accuracy of all three models. This suggests that recovering the high SNR self-distillation gradient is important for the model to extract more learning signal from the multi-token prediction objective, enough to \textbf{overtake NTP in zero-shot accuracy at 689M and 1.2B}, and in WikiText and LAMBADA perplexity at 1.2B.

\begin{table}[ht]
\centering
\small
\newcommand{\tsel}[1]{\,{\scriptsize\textcolor{gray}{(#1)}}}
\begin{tabular}{l|ccc|ccc|c|c}
\toprule
\multirow{2}{*}{\textbf{Model}}
 & \multicolumn{3}{c|}{\textbf{HumanEval}}
 & \multicolumn{3}{c|}{\textbf{MBPP}}
 & \textbf{GSM8K} & \textbf{MATH-500} \\
\cmidrule(lr){2-4} \cmidrule(lr){5-7} \cmidrule(lr){8-8} \cmidrule(lr){9-9}
 & pass@16 & pass@32 & pass@64 & pass@16 & pass@32 & pass@64 & acc & acc \\
\midrule
NTP
 & 28.89\tsel{T=0.5} & 33.77\tsel{T=0.6} & 39.02\tsel{T=0.6}
 & 47.83\tsel{T=0.5} & 53.65\tsel{T=0.5} & 58.80\tsel{T=0.5}
 & 47.69 & 24.80 \\
MTP
 & 31.70\tsel{T=0.5} & 36.37\tsel{T=0.5} & 40.85\tsel{T=0.7}
 & \textbf{48.28}\tsel{T=0.5} & 54.56\tsel{T=0.5} & 59.80\tsel{T=0.6}
 & 48.60 & 23.20 \\
\textbf{Soft-MTP}
 & \textbf{31.99}\tsel{T=0.5} & \textbf{37.26}\tsel{T=0.5} & \textbf{43.90}\tsel{T=0.6}
 & 47.88\tsel{T=0.5} & \textbf{54.59}\tsel{T=0.6} & \textbf{61.00}\tsel{T=0.7}
 & \textbf{49.66} & \textbf{25.00} \\
\bottomrule
\end{tabular}
\caption{\textbf{Math and code fine-tuning results}. Each 1.2B model is fine-tuned with the
NTP loss only, on the Python subset of Stack-Edu for code and on OpenMathInstruct-2 for math. For
code, we sample with top-$p=1.0$ and no top-$k$, sweep the temperature over
$\{0.1, 0.2, \ldots, 1.0\}$, and report the best pass@$k$, with the chosen temperature indicated in \textcolor{gray}{gray}.
For math, we evaluate 0-shot with greedy decoding. We \textbf{bold} the best result between MTP and Soft-MTP (NTP is shown for reference).}
\label{tab:finetune}
\end{table}

\paragraph{Math and code fine-tuning performance.} We next test how the pre-training objective affects downstream transfer to math and code reasoning. We discarded the MTP heads and fine-tuned each 1.2B model using the NTP loss alone. For math, we fine-tuned on 4.2B tokens from OpenMathInstruct-2 \citep{toshniwal2025openmathinstruct} and evaluated on the GSM8K \citep{cobbe2021training} and MATH-500 \citep{hendrycks2021measuring,lightman2023letsverifystepstep} benchmarks. For code, we fine-tuned on 21.0B tokens from the Python subset of Stack-Edu \citep{allal2025smollm2} and evaluated on the HumanEval \citep{chen2021evaluating} and MBPP \citep{austin2021program} benchmarks with the BigCode Evaluation Harness \citep{bigcode-evaluation-harness}. More details of the training hyperparameters and evaluation are provided in \cref{app:exp_hyperparam}. As shown in \cref{tab:finetune},  \textbf{Soft-MTP leads standard MTP in seven of the eight cells}, and is the only objective that never falls below the NTP baseline. On code, Soft-MTP's margin over standard MTP grows with the sample budget $k$, reaching $+3.1$ percentage points on HumanEval and $+1.2$ on MBPP at pass@64. On math, Soft-MTP leads standard MTP by $+1.1$ points on GSM8K and $+1.8$ points on MATH-500.

\begin{figure}[ht]
     \centering
     \begin{subfigure}[b]{0.245\textwidth}
         \centering
         \includegraphics[width=\textwidth, keepaspectratio]{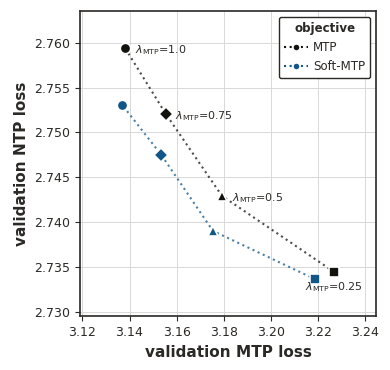}
         \caption{$\mathcal{L}_\text{NTP}$ vs. $\mathcal{L}_\text{MTP}$ across $\lambda_\text{MTP}$.}
         \label{fig:ntp_mtp_pareto}
     \end{subfigure}
     \hfill
     \begin{subfigure}[b]{0.245\textwidth}
         \centering
         \includegraphics[width=\textwidth, keepaspectratio]{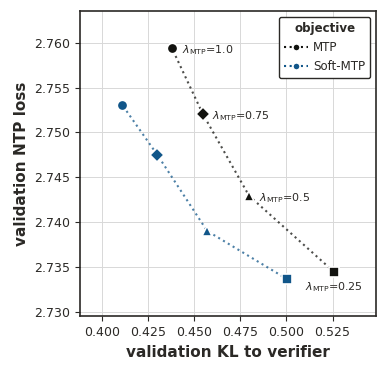}
         \caption{$\mathcal{L}_\text{NTP}$ vs. $D_\mathrm{KL}$ across $\lambda_\text{MTP}$.}
         \label{fig:ntp_kl_pareto}
     \end{subfigure}
     \hfill
     \begin{subfigure}[b]{0.245\textwidth}
         \centering
         \includegraphics[width=\textwidth, keepaspectratio]{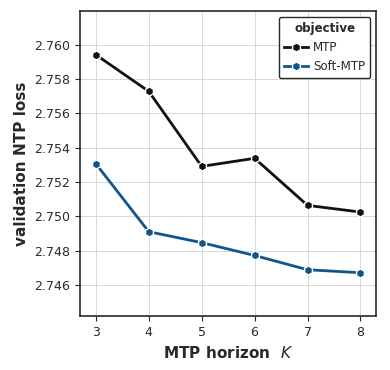}
         \caption{Horizon $K$ vs $\mathcal{L}_\text{NTP}$.}
         \label{fig:ntp_horizon}
     \end{subfigure}
     \hfill
     \begin{subfigure}[b]{0.245\textwidth}
         \centering
         \includegraphics[width=\textwidth, keepaspectratio]{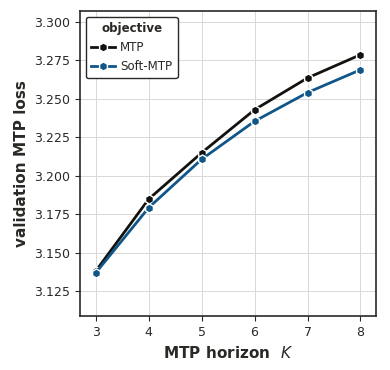}
         \caption{Horizon $K$ vs $\mathcal{L}_\text{MTP}$.}
         \label{fig:mtp_horizon}
     \end{subfigure}
     \caption{\textbf{Effect of MTP weight and prediction horizon} for 332M models trained on 7.3B tokens. Every axis is lower-better. (a-b) Varying MTP weight $\lambda_\text{MTP} \in \{0.25, 0.5, 0.75, 1.0\}$ at fixed horizon $K = 3$. Soft-MTP dominates the Pareto frontier for both plots. (c-d) Varying horizon $K \in \{3,4,5,6,7,8\}$ at fixed $\lambda_\text{MTP} = 1.0$. Soft-MTP leads on both axes at every $K$ shown.}
     \label{fig:effect_mtp_weight_horizon}
\end{figure}

\paragraph{Effect of MTP weight and horizon on training metrics.} To validate that Soft-MTP's advantage over standard MTP is robust to these hyperparameters, we sweep the MTP weight $\lambda_\text{MTP}$
and the horizon $K$ at the 332M scale, training each model on 7.3B FineWeb-Edu tokens. Lowering $\lambda_\text{MTP}$ trades MTP loss for NTP performance, but \cref{fig:ntp_mtp_pareto} shows that at every weight, \textbf{Soft-MTP reaches lower NTP and MTP loss than standard MTP}, so it Pareto-dominates standard MTP. The same holds for NTP loss against KL to the NTP verifier (see \cref{fig:ntp_kl_pareto}). Next, sweeping over MTP horizons while keeping $\lambda_\text{MTP}$ fixed at 1.0, \textbf{Soft-MTP also achieves lower NTP and MTP loss than standard MTP at every horizon} from $K=3$ to $K=8$
(see \cref{fig:ntp_horizon,fig:mtp_horizon}).

\begin{summarybox}
\textbf{Summary:} Soft-MTP attains lower NTP and MTP loss than standard MTP across weights $\lambda_\text{MTP}$ and horizons $K$. Its MTP heads sit closer to the NTP verifier in both KL and TV, thereby achieving faster speculative decoding. Soft-MTP also leads on zero-shot accuracy after pre-training and on both math and code reasoning after fine-tuning.
\end{summarybox}

\subsection{Soft-MTP is more efficient than KL-MTP} \label{subsection:soft_mtp_wins_kl_mtp}

\begin{table}[ht]
\centering
\small
\begin{tabular}{lccccc}
\toprule
 & \textbf{Matmul FLOPs} & \textbf{Throughput} & \textbf{MFU} & \multicolumn{2}{c}{\textbf{Peak memory} (GB)} \\
\cmidrule(lr){5-6}
\textbf{Objective} & (GFLOP/token) & (tok/s) & & naive CE & fused CE \\
\midrule
NTP               &  8.16 & 62{,}851 & 51.9\,\% &  98.3 &  95.5 \\
MTP               & 10.28 & 49{,}783 & 51.7\,\% & 120.7 & \textbf{107.0} \\
KL-MTP            & 10.43 & 46{,}990 & 49.6\,\% & 129.8 & 128.1 \\
\textbf{Soft-MTP} & 10.28 & 49{,}748 & 51.7\,\% & 121.0 & \textbf{107.0} \\
\bottomrule
\end{tabular}
\caption{Training cost by objective at the 1.2B model with $K=3$ MTP heads, measured on a single NVIDIA H200 GPU. Throughput and MFU are measured at the largest micro-batch size each objective fits. Peak memory is reported at a matched micro-batch of 19. \textbf{Soft-MTP's KL recovery is
free}: identical FLOPs, MFU and fused-CE peak memory to standard MTP, with throughput within $0.07\%$. \textbf{KL-MTP is worst on every axis}---lowest throughput and MFU, and the highest peak memory---and is the only objective that does not benefit much from using fused cross-entropy kernels.}
\label{tab:objective-cost}
\end{table}

\paragraph{KL-MTP is inefficient in compute and memory.} \Cref{tab:objective-cost} compares the training cost of each objective at 1.2B. Because the KL gradient--recovery trick reuses the cross-entropy gradients
that standard MTP already computes, Soft-MTP has the same matmul FLOPs, MFU and fused-CE peak
memory as standard MTP, and its throughput is within $0.07\%$. \textbf{KL-MTP is the most expensive on
every axis.} KL-MTP materializes both the teacher and student distributions over the vocabulary, which adds $1.5\%$ FLOPs, lowers throughput by $5.6\%$ relative to standard MTP, and raises peak memory by 21\,GB. It also gains little benefit from standard fused cross-entropy kernels because the KL objective needs dense distributions to be materialized.

\begin{table}[ht]
\centering
\small
\begin{tabular}{ll|cccc}
\toprule
 & \textbf{Metric} & \textbf{NTP} & \textbf{MTP} & \textbf{KL-MTP} & \textbf{Soft-MTP} \\
\midrule
\multirow{4}{*}{\rotatebox[origin=c]{90}{\scriptsize \textbf{LM}}}
 & FW-Edu ppl $\downarrow$    & 10.57 & 10.89 & \textbf{10.74} & \textbf{10.74} \\
 & WikiText ppl $\downarrow$  & 17.50 & 18.16 & 17.53 & \textbf{17.49} \\
 & LAMBADA ppl $\downarrow$   & 14.16 & 15.00 & 14.30 & \textbf{14.07} \\
 & Avg.\ accuracy $\uparrow$  & 56.75 & 56.80 & 57.22 & \textbf{57.31} \\
\midrule
\multirow{4}{*}{\rotatebox[origin=c]{90}{\scriptsize \textbf{Spec.\ dec.}}}
 & KL $\downarrow$            & --    & 0.384 & \textbf{0.363} & 0.364 \\
 & TV $\downarrow$            & --    & 0.260 & \textbf{0.252} & 0.253 \\
 & Acc.\ Len.\ $\uparrow$     & --    & 1.67  & \textbf{1.71}  & 1.70 \\
 & Speedup $\uparrow$         & --    & 1.74  & \textbf{1.76}  & \textbf{1.76} \\
\midrule
\multirow{2}{*}{\rotatebox[origin=c]{90}{\scriptsize \textbf{Math}}}
 & GSM8K $\uparrow$           & 47.69 & 48.60 & \textbf{50.34} & 49.66 \\
 & MATH-500 $\uparrow$        & 24.80 & 23.20 & \textbf{26.00} & 25.00 \\
\midrule
\multirow{6}{*}{\rotatebox[origin=c]{90}{\scriptsize \textbf{Code}}}
 & HumanEval pass@16 $\uparrow$ & 28.89 & 31.70 & 31.75 & \textbf{31.99} \\
 & HumanEval pass@32 $\uparrow$ & 33.77 & 36.37 & 36.41 & \textbf{37.26} \\
 & HumanEval pass@64 $\uparrow$ & 39.02 & 40.85 & 41.46 & \textbf{43.90} \\
 & MBPP pass@16 $\uparrow$      & 47.83 & \textbf{48.28} & 47.64 & 47.88 \\
 & MBPP pass@32 $\uparrow$      & 53.65 & 54.56 & 53.51 & \textbf{54.59} \\
 & MBPP pass@64 $\uparrow$      & 58.80 & 59.80 & 59.60 & \textbf{61.00} \\
\bottomrule
\end{tabular}
\caption{\textbf{Comparison against KL-MTP at 1.2B parameters.} We \textbf{bold} the best of
the three MTP objectives (NTP is shown for reference). \textbf{Language modeling (LM):} perplexity metrics and average
zero-shot accuracy over \textsc{lm-eval} tasks. \textbf{Speculative decoding (Spec.\ dec.):} self-speculative
decoding on FineWeb-Edu; acceptance length excludes the next-token prediction, which is
always accepted. \textbf{Fine-tuning (Math / Code):} performance after NTP-only fine-tuning on math and code.}
\label{tab:all_metrics_1536}
\end{table}

\paragraph{Soft-MTP matches or exceeds KL-MTP in most metrics.} \Cref{tab:all_metrics_1536} compares the three MTP objectives at 1.2B. Soft-MTP ties or wins on every perplexity metric, has higher average zero-shot accuracy, and beats KL-MTP on all six code fine-tuning metrics, while KL-MTP leads in math fine-tuning. In speculative decoding the two are neck and neck, with KL-MTP ahead by at most $0.01$ on FineWeb-Edu. This is expected because both objectives send the same KL gradient through the MTP heads and the shared trunk, so their MTP drafts should match the verifier about equally well. The two objectives differ only in the gradient used to update the output head, and we next explain why training it with cross-entropy, as Soft-MTP does, matters.

\paragraph{Why keep cross-entropy at the output head?} Note that the output head $W_\theta$ is shared in MTP; the same matrix maps both the NTP and the MTP hidden states to logits. As such, any gradient from
the MTP loss also shapes the NTP readout. Under Soft-MTP, this gradient is the MTP cross-entropy gradient which, like the NTP loss, fits the output head to the data distribution. Under KL-MTP, it is the self-distillation gradient, which fits the head to the model's own NTP predictions, and those predictions are themselves also computed by $W_\theta$. This creates a shortcut because the KL can be reduced not only by improving the MTP hidden states, but also by making the shared readout less sensitive to the directions in which the MTP and NTP hidden states differ. 
\begin{wrapfigure}{r}{0.4\textwidth}
\centering
\includegraphics[width=\linewidth]{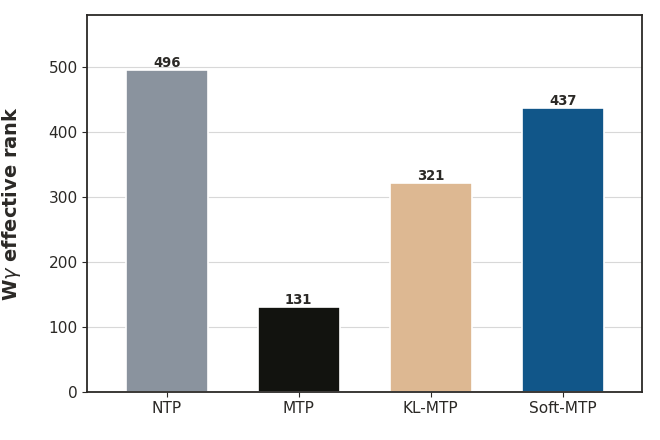}
\caption{\textbf{Effective rank of the logit map $W_\gamma$.} Soft-MTP stays closest
to the next-token readout.}
\label{fig:eff_rank}
\vspace{-1em}
\end{wrapfigure}
Cross-entropy has no such shortcut because its target does not depend on $W_\theta$. \Cref{fig:eff_rank} shows the consequence for the readout's geometry, measured by the effective
rank~\citep{roy2007effective} of the logit map $W_\gamma$, the unembedding combined with the
final RMSNorm gain $\gamma$ (we combine the two because the logits depend only on their
product). Standard MTP collapses the readout to about a quarter of the NTP baseline (131 vs.\
496) and KL-MTP to about two thirds (321), while Soft-MTP stays closest (437). The gradient routing differences between the objectives (see \cref{tab:gradient_routing}) allow us to pull the two effects apart. Soft-MTP and KL-MTP both send the KL gradient through the MTP heads and trunk and differ only at the output head, so the gap between them comes from keeping cross-entropy at the head. Soft-MTP and standard
MTP both use cross-entropy at the head and differ only in the trunk, so the much larger gap comes from removing the residual.  Soft-MTP combines the best of both worlds in terms of gradient routing, and it is also the objective with the strongest perplexity and downstream accuracy results.

\begin{summarybox}
\textbf{Summary:} Soft-MTP recovers the KL self-distillation gradient at the same cost as standard
MTP, whereas KL-MTP is less efficient in throughput and memory. Soft-MTP nonetheless
matches or exceeds KL-MTP on most metrics.
\end{summarybox}

\section{Discussion: Practical Relevance at Frontier Scale}
\label{section:practical_relevance}

MTP has been widely adopted in modern LLM training. Frontier model pre-training leverages many parallelism schemes due to tight memory budgets. Here we discuss the compatibility and relevance of Soft-MTP in frontier model pre-training.

\paragraph{Compatibility with parallelism.} The KL gradient--recovery trick needs only two
things where it is applied: the NTP hidden state $\hidden^{(0)}_{t+k}$ and the MTP hidden state
$\hidden^{(k)}_{t}$ on the same device, and a backward path from $\mathcal{L}_\text{NTP}$ to
$\hidden^{(k)}_{t}$. On its own, it adds no communication and no synchronization. Parallelism that shards parameters, optimizer state or
experts therefore leaves it untouched, e.g., data parallelism and ZeRO \citep{rajbhandari2020zero}, and expert parallelism \citep{lepikhin2021gshard} as used by DeepSeek-V3 \citep{deepseekai2024deepseekv3technicalreport} and Kimi K3 \citep{kimiteam2026kimik3openfrontier}. Because the gradient routing is built into the forward pass, the trick also needs only the single standard backward pass. In contrast, injecting $-\gres$ directly into $\hidden^{(k)}_t$, e.g., via a gradient hook, would have needed the NTP gradient first and hence two backward passes through the shared trunk, which distributed data parallelism \citep[DDP;][]{li2020pytorchdistributed} does not permit because a parameter cannot be marked ready for gradient synchronization twice in one iteration. Pipeline parallelism \citep{huang2019gpipe} only
requires the MTP modules to sit on the same stage as the output head, which DeepSeek-V3
already does by design so that its MTP module can share the output head. The only axis that needs extra communication is
sequence sharding, because our trick reads hidden states $k$ tokens ahead. This includes context parallelism, such as ring attention
\citep{liu2024ringattention}, and sequence parallelism \citep{korthikanti2022reducing}. Even then, the cost is a backward-pass exchange of at most $K$ hidden-state gradients at each shard boundary, independent of sequence length.

\paragraph{KL gradient--recovery trick is more compatible than explicit self-distillation.} An explicit KL needs the teacher's and the student's vocabulary-sized distributions at once, which standard fused cross-entropy kernels cannot compute without materializing the logits. This is especially costly for models trained without tensor parallelism, such as DeepSeek-V3 \citep{deepseekai2024deepseekv3technicalreport}, which therefore cannot shard these tensors along the vocabulary. A custom fused KL kernel can avoid this materialization, but must recompute the teacher logits that the NTP loss already computes, unless the NTP cross-entropy and all MTP KL terms are fused into a single kernel. The KL gradient--recovery trick instead carries the KL
gradient through the hidden state rather than the logits, so every head keeps its standard,
kernel-fused cross-entropy \citep{hsu2025ligerkernel,wijmans2025cut}, and Soft-MTP adds no vocabulary-sized tensors beyond those standard MTP already uses. Soft-MTP also matches, if not exceeds, direct KL minimization on most metrics as shown in \cref{subsection:soft_mtp_wins_kl_mtp}.

\paragraph{Future of MTP and extensions to the KL gradient--recovery trick.} MTP has spread rapidly
since DeepSeek-V3 adopted it in 2024~\citep{deepseekai2024deepseekv3technicalreport}, but its longevity in pre-training is uncertain. DeepSeek-V4.1-Flash \citep{deepseekai2026deepseekv41flashpushinglimitskv}
recently dropped MTP in favor of DSpark \citep{cheng2026dspark}, a speculative decoder trained after pre-training. Post-trained drafters such as Medusa \citep{cai2024medusa},
EAGLE \citep{li2024eagle1,li2024eagle2,li2026eagle}, DFlash \citep{chen2026dflash} and DSpark \citep{cheng2026dspark} are convenient, but it remains open how well they work when the base model was never trained to form predictive representations of future tokens, e.g., via MTP. Will MTP be retained for the superior downstream accuracy it provides over NTP-only pre-training \citep{gloeckle2024,deepseekai2024deepseekv3technicalreport,chen2026minimaxm2seriesminiactivations,nvidia2025nvidianemotron3efficient}, at the cost of minimal additional FLOPs? MTP has also proven useful in small-scale pre-training,
e.g., in Modded-NanoGPT \citep{modded_nanogpt_2024}, and in data-constrained settings, where it
improves sample efficiency, e.g., in NanoGPT Slowrun \citep{slowrun_2026}.
 Whatever the future of MTP, the KL gradient--recovery trick applies more broadly---whenever a teacher and a student share a linear output head, the gradient of the forward KL can be recovered from their cross-entropy gradients. Many post-training methods compute such a KL, for example MTP Boosting \citep{nvidia2026nemotron3ultraopen}, on-policy self-distillation
(OPSD; \citealp{zhao2026self}), and score centering for off-policy RL \citep{marek2026score}. Such use cases are expensive due to the KL computation overhead, which may necessitate top-$k$ logits approximation. Our KL gradient--recovery trick avoids that cost, with the caveat that only the MTP heads and shared trunk receive the exact KL gradient, whereas the shared output head $W_\theta$ is updated with cross-entropy. This is also suitable for methods that already freeze the output head under the KL loss, such as
Next-Latent Prediction \citep{teoh2026nextlatentpredictiontransformerslearn}, which only needs the KL gradient
with respect to the hidden states.

\section{Conclusion}

The standard MTP cross-entropy gradient can be decomposed into a KL self-distillation term, which pulls the MTP heads toward the model's own next-token distribution, and a residual, which is the next-token cross-entropy gradient against the data. The residual carries most of the MTP gradient, yet it has a low signal-to-noise ratio, works against next-token
learning, and becomes noisier with scale. Our KL gradient--recovery trick removes this residual and recovers the KL term exactly from the cross-entropy gradients that standard MTP already computes. This yields Soft-MTP, a one-line change to standard MTP that adds no parameters, has negligible
overhead, and remains compatible with fused cross-entropy kernels. Soft-MTP improves over standard MTP in
NTP and MTP loss, self-speculative decoding, zero-shot accuracy, and in math and code reasoning after fine-tuning, all at no extra cost.

\section*{Acknowledgements and Funding}
Experiments in this paper were mostly funded by Jayden, using $8\times$ RTX PRO 6000 GPUs rented from Vast.ai. The inspiration for this paper originally came from Jayden's attempt to amortize the computational cost of the KL objective in NextLat \citep{teoh2026nextlatentpredictiontransformerslearn}. 

We thank Alex L. Zhang, Alexi Gladstone, Daniel Suh, Edward S. Hu, John Langford, Khai Loong Aw, Michael K. Chen, Omri Homburger, Phillip Isola, Sophie L. Wang, and Stephen K. Yang for their valuable discussions and support. Jayden thanks Sophie L. Wang for helpful input on the design of the figures. Jayden also sincerely thanks Stephen K. Yang for sharing his Claude Max $20\times$ plan and for providing the support needed to push through this project.

\bibliography{ref} 

@article{radford2019language,
  title={Language models are unsupervised multitask learners},
  author={Radford, Alec and Wu, Jeffrey and Child, Rewon and Luan, David and Amodei, Dario and Sutskever, Ilya and others},
  journal={OpenAI blog},
  volume={1},
  number={8},
  pages={9},
  year={2019}
}

@inproceedings{toshniwal2025openmathinstruct,
  title={Openmathinstruct-2: Accelerating ai for math with massive open-source instruction data},
  author={Toshniwal, Shubham and Du, Wei and Moshkov, Ivan and Kisacanin, Branislav and Ayrapetyan, Alexan and Gitman, Igor},
  booktitle={International Conference on Learning Representations},
  volume={2025},
  pages={19243--19275},
  year={2025}
}

@article{allal2025smollm2,
  title={SmolLM2: When Smol Goes Big--Data-Centric Training of a Small Language Model},
  author={Allal, Loubna Ben and Lozhkov, Anton and Bakouch, Elie and Bl{\'a}zquez, Gabriel Mart{\'\i}n and Penedo, Guilherme and Tunstall, Lewis and Marafioti, Andr{\'e}s and Kydl{\'\i}{\v{c}}ek, Hynek and Lajar{\'\i}n, Agust{\'\i}n Piqueres and Srivastav, Vaibhav and others},
  journal={arXiv preprint arXiv:2502.02737},
  year={2025}
}

@misc{sutskever2023nexttoken,
  author       = {Ilya Sutskever},
  title        = {Why Next-Token Prediction Is Enough for AGI},
  year         = {2023},
  month        = dec,
  howpublished = {YouTube video},
  publisher    = {Dwarkesh Patel},
  url          = {https://www.youtube.com/watch?v=YEUclZdj_Sc},
  note         = {Published December 13, 2023}
}

@article{teoh2026nextlatentpredictiontransformerslearn,
  title={Next-latent prediction transformers learn compact world models},
  author={Teoh, Jayden and Tomar, Manan and Ahn, Kwangjun and Hu, Edward S. and Pearce, Tim and Sharma, Pratyusha and Krishnamurthy, Akshay and Islam, Riashat and Lamb, Alex and Langford, John},
  journal={arXiv preprint arXiv:2511.05963},
  year={2025}
}

@inproceedings{Mahajan2026,
 author = {Mahajan, Divyat and Goyal, Sachin and Youbi Idrissi, Badr and Pezeshki, Mohammad and Mitliagkas, Ioannis and Lopez-Paz, David and Ahuja, Kartik},
 booktitle = {International Conference on Learning Representations},
 editor = {C. Vondrick and B. Hariharan and C. Raffel and L. Pinto and D. Yang and A. Faust},
 pages = {64342--64364},
 title = {Beyond Multi-Token Prediction: Pretraining LLMs with Future Summaries},
 url = {https://proceedings.iclr.cc/paper_files/paper/2026/file/693d7541bd31cca2dcb780abfd8d204c-Paper-Conference.pdf},
 volume = {2026},
 year = {2026}
}

@inproceedings{
zuhri2026predicting,
title={Predicting the Order of Upcoming Tokens Improves Language Modeling},
author={Zayd Muhammad Kawakibi Zuhri and Erland Hilman Fuadi and Alham Fikri Aji},
booktitle={Forty-third International Conference on Machine Learning},
year={2026},
url={https://openreview.net/forum?id=wmROqgop27}
}

@inproceedings{qi2020prophetnet,
  title={Prophetnet: Predicting future n-gram for sequence-to-sequence pre-training},
  author={Qi, Weizhen and Yan, Yu and Gong, Yeyun and Liu, Dayiheng and Duan, Nan and Chen, Jiusheng and Zhang, Ruofei and Zhou, Ming},
  booktitle={Proceedings of the 2020 Conference on Empirical Methods in Natural Language Processing: Findings},
  pages={2401--2410},
  year={2020}
}

@article{deepseekai2024deepseekv3technicalreport,
  title={DeepSeek-V3 Technical Report},
  author={DeepSeek-AI},
  journal={arXiv preprint arXiv:2412.19437},
  year={2024}
}

@article{deepseekai2026deepseekv41flashpushinglimitskv,
  title={DeepSeek-V4.1-Flash: Pushing the Limits of KV Cache Compression},
  author={DeepSeek-AI},
  journal={arXiv preprint arXiv:2609.19969},
  year={2026}
}

@article{cheng2026dspark,
  title={DSpark: Confidence-Scheduled Speculative Decoding with Semi-Autoregressive Generation},
  author={Cheng, Xin and Yu, Xingkai and Shao, Chenze and Li, Jiashi and Xiong, Yunfan and Qian, Yi and Zhu, Jiaqi and Ma, Shirong and Zhang, Xiaokang and Ye, Jiasheng and others},
  journal={arXiv preprint arXiv:2607.05147},
  year={2026}
}

@article{chen2026dflash,
  title={Dflash: Block diffusion for flash speculative decoding},
  author={Chen, Jian and Liang, Yesheng and Liu, Zhijian},
  journal={arXiv preprint arXiv:2602.06036},
  year={2026}
}

@article{zhao2026self,
  title={Self-distilled reasoner: On-policy self-distillation for large language models},
  author={Zhao, Siyan and Xie, Zhihui and Liu, Mengchen and Huang, Jing and Pang, Guan and Chen, Feiyu and Grover, Aditya},
  journal={arXiv preprint arXiv:2601.18734},
  year={2026}
}

@article{marek2026score,
  title={Score centering stabilizes off-policy reinforcement learning},
  author={Marek, Martin and Ryabinin, Max},
  journal={arXiv preprint arXiv:2609.20807},
  year={2026}
}

@misc{modded_nanogpt_2024,
  author       = {Keller Jordan and Jeremy Bernstein and Brendan Rappazzo and
                  {@fernbear.bsky.social} and Boza Vlado and You Jiacheng and
                  Franz Cesista and Braden Koszarsky and {@Grad62304977}},
  title        = {modded-nanogpt: Speedrunning the {NanoGPT} baseline},
  year         = {2024},
  url          = {https://github.com/KellerJordan/modded-nanogpt},
  note         = {Multi-token prediction added in pull request \#178}
}

@misc{slowrun_2026,
  author       = {Akshay Vegesna and Samip Dahal and Chinmay Karkar and
                  Bishwas Mandal and Shmuel Berman and Zhiwei Xu},
  title        = {Slowrun: Language modeling with infinite compute, fixed data},
  year         = {2026},
  url          = {https://github.com/qlabs-eng/slowrun}
}

@article{li2026eagle,
  title={Eagle-3: Scaling up inference acceleration of large language models via training-time test},
  author={Li, Yuhui and Wei, Fangyun and Zhang, Chao and Zhang, Hongyang},
  journal={Advances in Neural Information Processing Systems},
  volume={38},
  pages={136737--136756},
  year={2026}
}

@article{qwen3technicalreport,
  title={Qwen3 Technical Report},
  author={{Qwen Team}},
  journal={arXiv preprint arXiv:2505.09388},
  year={2025}
}

@article{deepseekai2026deepseekv4,
  title={DeepSeek-V4: Towards Highly Efficient Million-Token Context Intelligence},
  author={DeepSeek-AI},
  journal={arXiv preprint arXiv:2606.19348},
  year={2026}
}

@article{meituan2025longcatflashtechnicalreport,
  title={LongCat-Flash Technical Report},
  author={Meituan LongCat},
  journal={arXiv preprint arXiv:2509.01322},
  year={2025}
}

@article{nvidia2025nvidianemotron3efficient,
  title={NVIDIA Nemotron 3: Efficient and Open Intelligence},
  author={NVIDIA},
  journal={arXiv preprint arXiv:2512.20856},
  year={2025}
}

@article{nvidia2026nemotron3ultraopen,
  title={Nemotron 3 Ultra: Open, Efficient Mixture-of-Experts Hybrid Mamba-Transformer Model for Agentic Reasoning},
  author={NVIDIA},
  journal={arXiv preprint arXiv:2606.15007},
  year={2026}
}

@article{coreteam2026mimov2flashtechnicalreport,
  title={MiMo-V2-Flash Technical Report},
  author={LLM-Core Xiaomi},
  journal={arXiv preprint arXiv:2601.02780},
  year={2026}
}

@article{chen2026minimaxm2seriesminiactivations,
  title={The MiniMax-M2 Series: Mini Activations Unleashing Max Real-World Intelligence},
  author={MiniMax},
  journal={arXiv preprint arXiv:2605.26494},
  year={2026}
}

@article{kimiteam2026kimik3openfrontier,
  title={Kimi K3: Open Frontier Intelligence},
  author={{Kimi Team}},
  journal={arXiv preprint arXiv:2607.24653},
  year={2026}
}

@article{5team2025glm45agenticreasoningcoding,
  title={GLM-4.5: Agentic, Reasoning, and Coding (ARC) Foundation Models},
  author={{GLM Team}},
  journal={arXiv preprint arXiv:2508.06471},
  year={2025}
}

@article{glm5team2026glm5vibecodingagentic,
  title={GLM-5: from Vibe Coding to Agentic Engineering},
  author={GLM-5-Team},
  journal={arXiv preprint arXiv:2602.15763},
  year={2026}
}

@article{lingteam2025activationboostedscalinggeneral,
  title={Every Activation Boosted: Scaling General Reasoner to 1 Trillion Open Language Foundation},
  author={{Ling Team}},
  journal={arXiv preprint arXiv:2510.22115},
  year={2025}
}

@article{bengio2013estimatingpropagatinggradientsstochastic,
  title={Estimating or Propagating Gradients Through Stochastic Neurons for Conditional Computation},
  author={Yoshua Bengio and Nicholas Léonard and Aaron Courville},
  journal={arXiv preprint arXiv:1308.3432},
  year={2013}
}

@inproceedings{vqvae2017,
author = {van den Oord, Aaron and Vinyals, Oriol and Kavukcuoglu, Koray},
title = {Neural discrete representation learning},
year = {2017},
isbn = {9781510860964},
publisher = {Curran Associates Inc.},
address = {Red Hook, NY, USA},
booktitle = {Proceedings of the 31st International Conference on Neural Information Processing Systems},
pages = {6309–6318},
numpages = {10},
location = {Long Beach, California, USA},
series = {NIPS'17}
}

@misc{chang2026klcachetrick,
  title = {KL Cache Trick for Full-Vocab Distillation},
  author = {Jonathan Chang and {gpt-5.5-xhigh}},
  year = {2026},
  url = {https://jonathanc.net/blog/kl-cache-trick}
}

@article{kirchenbauer2026multitokenpredictionselfdistillation,
  title={Multi-Token Prediction via Self-Distillation},
  author={John Kirchenbauer and Abhimanyu Hans and Brian Bartoldson and Micah Goldblum and Ashwinee Panda and Tom Goldstein},
  journal={arXiv preprint arXiv:2602.06019},
  year={2026}
}

@article{zhao2026selfdistillationmultitokenprediction,
  title={Self-Distillation for Multi-Token Prediction},
  author={Guoliang Zhao and Ruobing Xie and An Wang and Shuaipeng Li and Huaibing Xie and Xingwu Sun},
  journal={arXiv preprint arXiv:2603.23911},
  year={2026}
}

@article{mccandlish2018,
  title={An Empirical Model of Large-Batch Training},
  author={Sam McCandlish and Jared Kaplan and Dario Amodei and OpenAI Dota Team},
  journal={arXiv preprint arXiv:1812.06162},
  year={2018}
}

@article{touvron2023llama,
  title={Llama: Open and efficient foundation language models},
  author={Touvron, Hugo and Lavril, Thibaut and Izacard, Gautier and Martinet, Xavier and Lachaux, Marie-Anne and Lacroix, Timoth{\'e}e and Rozi{\`e}re, Baptiste and Goyal, Naman and Hambro, Eric and Azhar, Faisal and others},
  journal={arXiv preprint arXiv:2302.13971},
  year={2023}
}

@article{hoffmann2022training,
  title={Training compute-optimal large language models},
  author={Hoffmann, Jordan and Borgeaud, Sebastian and Mensch, Arthur and Buchatskaya, Elena and Cai, Trevor and Rutherford, Eliza and Casas, Diego de Las and Hendricks, Lisa Anne and Welbl, Johannes and Clark, Aidan and others},
  journal={arXiv preprint arXiv:2203.15556},
  year={2022}
}

@article{cobbe2021training,
  title={Training verifiers to solve math word problems},
  author={Cobbe, Karl and Kosaraju, Vineet and Bavarian, Mohammad and Chen, Mark and Jun, Heewoo and Kaiser, Lukasz and Plappert, Matthias and Tworek, Jerry and Hilton, Jacob and Nakano, Reiichiro and others},
  journal={arXiv preprint arXiv:2110.14168},
  year={2021}
}

@inproceedings{
hendrycks2021measuring,
title={Measuring Mathematical Problem Solving With the {MATH} Dataset},
author={Dan Hendrycks and Collin Burns and Saurav Kadavath and Akul Arora and Steven Basart and Eric Tang and Dawn Song and Jacob Steinhardt},
booktitle={Thirty-fifth Conference on Neural Information Processing Systems Datasets and Benchmarks Track (Round 2)},
year={2021},
url={https://openreview.net/forum?id=7Bywt2mQsCe}
}

@article{chen2021evaluating,
  title={Evaluating large language models trained on code},
  author={Chen, Mark and Tworek, Jerry and Jun, Heewoo and Yuan, Qiming and Pinto, Henrique Ponde De Oliveira and Kaplan, Jared and Edwards, Harri and Burda, Yuri and Joseph, Nicholas and Brockman, Greg and others},
  journal={arXiv preprint arXiv:2107.03374},
  year={2021}
}

@article{austin2021program,
  title={Program synthesis with large language models},
  author={Austin, Jacob and Odena, Augustus and Nye, Maxwell and Bosma, Maarten and Michalewski, Henryk and Dohan, David and Jiang, Ellen and Cai, Carrie and Terry, Michael and Le, Quoc and others},
  journal={arXiv preprint arXiv:2108.07732},
  year={2021}
}

@misc{bigcode-evaluation-harness,
  author       = {Ben Allal, Loubna and
                  Muennighoff, Niklas and
                  Kumar Umapathi, Logesh and
                  Lipkin, Ben and
                  von Werra, Leandro},
  title = {A framework for the evaluation of code generation models},
  publisher = {GitHub},
  journal = {GitHub repository},
  howpublished = {\url{https://github.com/bigcode-project/bigcode-evaluation-harness}},
  year = 2022,
}

@article{lightman2023letsverifystepstep,
  title={Let's Verify Step by Step},
  author={Hunter Lightman and Vineet Kosaraju and Yura Burda and Harri Edwards and Bowen Baker and Teddy Lee and Jan Leike and John Schulman and Ilya Sutskever and Karl Cobbe},
  journal={arXiv preprint arXiv:2305.20050},
  year={2023}
}

@article{hu2025the,
  title={The Belief State Transformer},
  author={Edward S. Hu and Kwangjun Ahn and Qinghua Liu and Haoran Xu and Manan Tomar and Ada Langford and Jayden Teoh and Bryon Xu and David Yan and Dinesh Jayaraman and Alex Lamb and John Langford},
  journal={arXiv preprint arXiv:2410.23506},
  year={2025}
}

@inproceedings{
huang2026how,
title={How Transformers Learn to Plan via Multi-Token Prediction},
author={Jianhao Huang and Zhanpeng Zhou and Renqiu Xia and Baharan Mirzasoleiman and Weijie J Su and Wei Huang},
booktitle={Third Conference on Language Modeling},
year={2026},
url={https://openreview.net/forum?id=1uj8eRv3aJ}
}

@inproceedings{
liu2025lmtp,
title={L-{MTP}: Leap Multi-Token Prediction Beyond Adjacent Context for Large Language Models},
author={Xiaohao Liu and Xiaobo Xia and Weixiang Zhao and Manyi Zhang and Xianzhi Yu and Xiu Su and Shuo Yang and See-Kiong Ng and Tat-Seng Chua},
booktitle={The Thirty-ninth Annual Conference on Neural Information Processing Systems},
year={2025},
url={https://openreview.net/forum?id=0VDmWjW456}
}

@inproceedings{
dao2024transformers,
title={Transformers are {SSM}s: Generalized Models and Efficient Algorithms Through Structured State Space Duality},
author={Tri Dao and Albert Gu},
booktitle={Forty-first International Conference on Machine Learning},
year={2024},
url={https://openreview.net/forum?id=ztn8FCR1td}
}

@inproceedings{rajbhandari2020zero,
author = {Rajbhandari, Samyam and Rasley, Jeff and Ruwase, Olatunji and He, Yuxiong},
title = {ZeRO: memory optimizations toward training trillion parameter models},
year = {2020},
isbn = {9781728199986},
publisher = {IEEE Press},
booktitle = {Proceedings of the International Conference for High Performance Computing, Networking, Storage and Analysis},
articleno = {20},
numpages = {16},
location = {Atlanta, Georgia},
series = {SC '20}
}

@inproceedings{
lepikhin2021gshard,
title={GShard: Scaling Giant Models with Conditional Computation and Automatic Sharding}, 
author={Dmitry Lepikhin and HyoukJoong Lee and Yuanzhong Xu and Dehao Chen and Orhan Firat and Yanping Huang and Maxim Krikun and Noam Shazeer and Zhifeng Chen},
booktitle={International Conference on Learning Representations},
year={2021},
url={https://openreview.net/forum?id=qrwe7XHTmYb}
}

@article{huang2019gpipe,
  title={Gpipe: Efficient training of giant neural networks using pipeline parallelism},
  author={Huang, Yanping and Cheng, Youlong and Bapna, Ankur and Firat, Orhan and Chen, Dehao and Chen, Mia and Lee, HyoukJoong and Ngiam, Jiquan and Le, Quoc V and Wu, Yonghui and others},
  journal={Advances in neural information processing systems},
  volume={32},
  year={2019}
}

@inproceedings{
liu2024ringattention,
title={RingAttention with Blockwise Transformers for Near-Infinite Context},
author={Hao Liu and Matei Zaharia and Pieter Abbeel},
booktitle={The Twelfth International Conference on Learning Representations},
year={2024},
url={https://openreview.net/forum?id=WsRHpHH4s0}
}

@article{korthikanti2022reducing,
  title={Reducing Activation Recomputation in Large Transformer Models},
  author={Vijay Korthikanti and Jared Casper and Sangkug Lym and Lawrence McAfee and Michael Andersch and Mohammad Shoeybi and Bryan Catanzaro},
  journal={arXiv preprint arXiv:2205.05198},
  year={2022}
}

@article{hinton2015distilling,
  title={Distilling the knowledge in a neural network},
  author={Hinton, Geoffrey and Vinyals, Oriol and Dean, Jeff},
  journal={arXiv preprint arXiv:1503.02531},
  year={2015}
}

@article{ahn2025jtp,
  title={Efficient Joint Prediction of Multiple Future Tokens},
  author={Kwangjun Ahn and Alex Lamb and John Langford},
  journal={arXiv preprint arXiv:2503.21801},
  year={2025}
}

@InProceedings{bachmann24a,
  title = 	 {The Pitfalls of Next-Token Prediction},
  author =       {Bachmann, Gregor and Nagarajan, Vaishnavh},
  booktitle = 	 {Proceedings of the 41st International Conference on Machine Learning},
  pages = 	 {2296--2318},
  year = 	 {2024},
  editor = 	 {Salakhutdinov, Ruslan and Kolter, Zico and Heller, Katherine and Weller, Adrian and Oliver, Nuria and Scarlett, Jonathan and Berkenkamp, Felix},
  volume = 	 {235},
  series = 	 {Proceedings of Machine Learning Research},
  month = 	 {21--27 Jul},
  publisher =    {PMLR},
  url = 	 {https://proceedings.mlr.press/v235/bachmann24a.html},
}

@inproceedings{gloeckle2024,
author = {Gloeckle, Fabian and Idrissi, Badr Youbi and Rozi\`{e}re, Baptiste and Lopez-Paz, David and Synnaeve, Gabriel},
title = {Better \& faster large language models via multi-token prediction},
year = {2024},
publisher = {JMLR.org},
booktitle = {Proceedings of the 41st International Conference on Machine Learning},
articleno = {629},
numpages = {29},
location = {Vienna, Austria},
series = {ICML'24}
}

@article{nagarajan2025roll,
  title={Roll the dice \& look before you leap: Going beyond the creative limits of next-token prediction},
  author={Nagarajan, Vaishnavh and Wu, Chen Henry and Ding, Charles and Raghunathan, Aditi},
  journal={arXiv preprint arXiv:2504.15266},
  year={2025}
}

@inproceedings{roy2007effective,
  title={The effective rank: A measure of effective dimensionality},
  author={Roy, Olivier and Vetterli, Martin},
  booktitle={2007 15th European signal processing conference},
  pages={606--610},
  year={2007},
  organization={IEEE}
}

@article{loshchilov2019decoupledweightdecayregularization,
  title={Decoupled Weight Decay Regularization},
  author={Ilya Loshchilov and Frank Hutter},
  journal={arXiv preprint arXiv:1711.05101},
  year={2019}
}

@inproceedings{leviathan2022fast,
author = {Leviathan, Yaniv and Kalman, Matan and Matias, Yossi},
title = {Fast inference from transformers via speculative decoding},
year = {2023},
publisher = {JMLR.org},
booktitle = {Proceedings of the 40th International Conference on Machine Learning},
articleno = {795},
numpages = {13},
location = {Honolulu, Hawaii, USA},
series = {ICML'23}
}

@article{chen2023acceleratinglargelanguagemodel,
  title={Accelerating Large Language Model Decoding with Speculative Sampling},
  author={Charlie Chen and Sebastian Borgeaud and Geoffrey Irving and Jean-Baptiste Lespiau and Laurent Sifre and John Jumper},
  journal={arXiv preprint arXiv:2302.01318},
  year={2023}
}

@article{stern2018blockwise,
  title={Blockwise parallel decoding for deep autoregressive models},
  author={Stern, Mitchell and Shazeer, Noam and Uszkoreit, Jakob},
  journal={Advances in Neural Information Processing Systems},
  volume={31},
  year={2018}
}

@article{li2024eagle1,
  title={Eagle: Speculative sampling requires rethinking feature uncertainty},
  author={Li, Yuhui and Wei, Fangyun and Zhang, Chao and Zhang, Hongyang},
  journal={arXiv preprint arXiv:2401.15077},
  year={2024}
}

@inproceedings{li2024eagle2,
  title={Eagle-2: Faster inference of language models with dynamic draft trees},
  author={Li, Yuhui and Wei, Fangyun and Zhang, Chao and Zhang, Hongyang},
  booktitle={Proceedings of the 2024 conference on empirical methods in natural language processing},
  pages={7421--7432},
  year={2024}
}

@article{cai2024medusa,
  title={Medusa: Simple llm inference acceleration framework with multiple decoding heads},
  author={Cai, Tianle and Li, Yuhong and Geng, Zhengyang and Peng, Hongwu and Lee, Jason D and Chen, Deming and Dao, Tri},
  journal={arXiv preprint arXiv:2401.10774},
  year={2024}
}

@article{penedo2024fineweb,
  title={The fineweb datasets: Decanting the web for the finest text data at scale},
  author={Penedo, Guilherme and Kydl{\'\i}{\v{c}}ek, Hynek and Lozhkov, Anton and Mitchell, Margaret and Raffel, Colin and Von Werra, Leandro and Wolf, Thomas and others},
  journal={Advances in Neural Information Processing Systems},
  volume={37},
  pages={30811--30849},
  year={2024}
}

@misc{eval-harness,
  author       = {Gao, Leo and Tow, Jonathan and Abbasi, Baber and Biderman, Stella and Black, Sid and DiPofi, Anthony and Foster, Charles and Golding, Laurence and Hsu, Jeffrey and Le Noac'h, Alain and Li, Haonan and McDonell, Kyle and Muennighoff, Niklas and Ociepa, Chris and Phang, Jason and Reynolds, Laria and Schoelkopf, Hailey and Skowron, Aviya and Sutawika, Lintang and Tang, Eric and Thite, Anish and Wang, Ben and Wang, Kevin and Zou, Andy},
  title        = {The Language Model Evaluation Harness},
  month        = 07,
  year         = 2024,
  publisher    = {Zenodo},
  version      = {v0.4.3},
  doi          = {10.5281/zenodo.12608602},
  url          = {https://zenodo.org/records/12608602}
}

@misc{jordan2024muon,
  author       = {Keller Jordan and Yuchen Jin and Vlado Boza and You Jiacheng and
                  Franz Cesista and Laker Newhouse and Jeremy Bernstein},
  title        = {Muon: An optimizer for hidden layers in neural networks},
  year         = {2024},
  url          = {https://kellerjordan.github.io/posts/muon/}
}

@article{liu2025muon,
  title={Muon is scalable for llm training},
  author={Liu, Jingyuan and Su, Jianlin and Yao, Xingcheng and Jiang, Zhejun and Lai, Guokun and Du, Yulun and Qin, Yidao and Xu, Weixin and Lu, Enzhe and Yan, Junjie and others},
  journal={arXiv preprint arXiv:2502.16982},
  year={2025}
}

@article{li2026breakingentropyboundsaccelerating,
  title={Breaking Entropy Bounds: Accelerating RL Training via MTP with Rejection Sampling},
  author={Yucheng Li and Huiqiang Jiang and Yang Xu and Jianxin Yang and Yi Zhang and Yizhong Cao and Yuhao Shen and Fan Zhou and Rui Men and Jianwei Zhang and An Yang and Bowen Yu and Bo Zheng and Fei Huang and Junyang Lin and Dayiheng Liu and Jingren Zhou},
  journal={arXiv preprint arXiv:2606.12370},
  year={2026}
}

@inproceedings{
samarin2026lk,
title={{LK} Losses: Direct Acceptance Rate Optimization for Speculative Decoding},
author={Alexander Samarin and Sergei Krutikov and Anton Shevtsov and Sergei Skvortsov and Filipp Fisin and Alexander Golubev},
booktitle={Forty-third International Conference on Machine Learning},
year={2026},
url={https://openreview.net/forum?id=ZqnVidXtxV}
}

@inproceedings{
hsu2025ligerkernel,
title={Liger-Kernel: Efficient Triton Kernels for {LLM} Training},
author={Pin-Lun Hsu and Yun Dai and Vignesh Kothapalli and Qingquan Song and Shao Tang and Siyu Zhu and Steven Shimizu and Shivam Sahni and Haowen Ning and Yanning Chen and Zhipeng Wang},
booktitle={Championing Open-source DEvelopment in ML Workshop @ ICML25},
year={2025},
url={https://openreview.net/forum?id=36SjAIT42G}
}

@inproceedings{
wijmans2025cut,
title={Cut Your Losses in Large-Vocabulary Language Models},
author={Erik Wijmans and Brody Huval and Alexander Hertzberg and Vladlen Koltun and Philipp Kraehenbuehl},
booktitle={The Thirteenth International Conference on Learning Representations},
year={2025},
url={https://openreview.net/forum?id=E4Fk3YuG56}
}

@article{iso2026rlacceleration,
  title={Accelerating RL Post-Training Rollouts via System-Integrated Speculative Decoding},
  author={Hayate Iso and Tiyasa Mitra and Sudipta Mondal and Rasoul Shafipour and Venmugil Elango and Terry Kong and Yuki Huang and Seonjin Na and Izzy Putterman and Benjamin Chislett and Maor Ashkenazi and Joseph Guman and Gerald Shen and Tugrul Konuk and Ashwath Aithal and Ritika Borkar and Ran Zilberstein and Bita Rouhani},
  journal={arXiv preprint arXiv:2604.26779},
  year={2026}
}

@article{li2020pytorchdistributed,
  title={{PyTorch} Distributed: Experiences on Accelerating Data Parallel Training},
  author={Li, Shen and Zhao, Yanli and Varma, Rohan and Salpekar, Omkar and Noordhuis, Pieter and Li, Teng and Paszke, Adam and Smith, Jeff and Vaughan, Brian and Damania, Pritam and Chintala, Soumith},
  journal={Proceedings of the VLDB Endowment},
  volume={13},
  number={12},
  pages={3005--3018},
  year={2020}
}
\bibliographystyle{plainnat}

\clearpage
\appendix

\section{Full Soft-MTP PyTorch Implementation}
\label{app:full_pseudocode}

\begin{algorithm}[htbp]
\caption{Full Soft-MTP implementation in PyTorch syntax, for DeepSeek-style sequential MTP.
The scale $s_k$ rescales each routed gradient to match the corresponding head's own
cross-entropy gradient (\cref{app:loss_weighting}); the highlighted lines are the
only changes from standard MTP.}
\label{alg:full_soft_mtp_code}
{\renewcommand{\codesize}{\footnotesize}
\begin{lstlisting}[style=minimalcode, basicstyle=\ttfamily\codesize\color{codetext}]
# x, y       : [B, T] input tokens and pre-shifted labels; y[:, t] is the
#              target of position t.
# K          : number of MTP heads.
# lambda_mtp : auxiliary MTP weight.
# ce         : cross-entropy, mean-reduced over the targets it is given.
#
# MTP head k at position t and the NTP head at position t+k predict the
# SAME token y[:, t+k]; that shared label is what cancels the label term
# and leaves the KL gradient. Each MTP head has one fewer target than the 
# head before it: head k has T-k targets, y[:, k:].

h_prev  = shared_trunk(x)
h0 = norm[0](h_prev) # NTP hidden states are directly from trunk, [B, T, d]
T  = y.size(1)

loss_mtp = 0
for k in range(1, K+1):
    yk = y[:, k:]  # [B, T-k] k-th MTP head targets
    h_prev = mtp_head[k-1](h_prev[:, :T-k], x[:, k:]) # DeepSeek's sequential MTP
    hk = norm[k](h_prev) 

    # hard CE: hk gets lambda_mtp/(K(T-k)) * CE gradient
    loss_mtp += ce(lm_head, hk, yk)

    # straight-through gradient routing
    # s_k undoes the NTP CE's 1/T mean and re-applies k-th MTP head's own
    # 1/(T-k) mean and the 1/K head average, so the routed gradient
    # lands at exactly the scale the MTP head's own CE gradient uses.
    (*!\hlcode{s\_k = lambda\_mtp * T / (K * (T - k))}!*)

    # h0 is numerically unchanged; only its backward changes
    (*!\hlcode{h0[:, k:] = h0[:, k:] + s\_k * (hk.detach() - hk)}!*)

# NTP loss and the gradients reaching the trunk through h0 are unchanged
loss_ntp = ce(lm_head, h0, y)

# backward: hk[t] gets -s_k * (NTP CE grad at h0[t+k]); its one-hot part
# cancels the one in hk's own CE grad, leaving the KL gradient
# lambda_mtp / (K(T-k)) * W^T (p_k[t] - p_0[t+k])
(loss_ntp + lambda_mtp/K * loss_mtp).backward()
\end{lstlisting}
}
\end{algorithm}

\section{Extra Derivations for MTP Gradients}
\label{app:mtp_grad}

Throughout, $W_\theta \in \mathbb{R}^{V\times d}$ is the shared linear output head, $\hidden \in
\mathbb{R}^d$ denotes the hidden state fed into it (after any final normalization), $z = W_\theta
\hidden$ are the logits, and $p = \mathrm{softmax}(z)$. Targets $y \in \mathbb{R}^V$ may be one-hot
vectors or full distributions, with $\sum_v y_v = 1$.

\subsection{Hard MTP gradient}
We derive \cref{eq:hard_mtp_grad},
\begin{align*}
    g_\text{hard} &\coloneqq \nabla_{\hidden_{t}^{(k)}} \mathrm{CE}\left(x_{t+k+1}, p_t^{(k)}\right) \\
    &= W_\theta^\intercal\left(p_t^{(k)} - x_{t+k+1}\right). \tag{\ref*{eq:hard_mtp_grad}}
\end{align*}
First, the cross-entropy gradient with respect to the logits. Since $\log p_v = z_v - \log\sum_u
e^{z_u}$ and $\sum_v y_v = 1$,
\begin{align*}
    \mathrm{CE}(y, p) = -\sum_v y_v \log p_v = -y^\intercal z + \log\sum_u e^{z_u},
\end{align*}
and differentiating gives $\partial\, \mathrm{CE}/\partial z_v = -y_v + e^{z_v}/\sum_u e^{z_u} = p_v - y_v$, i.e.
\begin{align}
    \nabla_z \mathrm{CE}\left(y, \mathrm{softmax}(z)\right) = p - y. \label{eq:app_ce_logit_grad}
\end{align}
The $k$-th MTP logits are $z_t^{(k)} = W_\theta \hidden_t^{(k)}$, so $\partial z_t^{(k)} / \partial
\hidden_t^{(k)} = W_\theta$. By the chain rule and \cref{eq:app_ce_logit_grad} with $y = x_{t+k+1}$,
\begin{align*}
    \nabla_{\hidden_t^{(k)}} \mathrm{CE}\left(x_{t+k+1}, p_t^{(k)}\right)
    = \left(\frac{\partial z_t^{(k)}}{\partial \hidden_t^{(k)}}\right)^{\!\intercal} \nabla_{z_t^{(k)}} \mathrm{CE}
    = W_\theta^\intercal\left(p_t^{(k)} - x_{t+k+1}\right).
\end{align*}

\subsection{KL self-distillation gradient}
We derive \cref{eq:kl_grad},
\begin{align*}
    \nabla_{\hidden_{t}^{(k)}}D_\mathrm{KL}\left(\stopgrad(q_t^{(k)}) \| p_t^{(k)}\right) = W_\theta^\intercal \left(p_t^{(k)} - q_t^{(k)}\right) = \gsoft. \tag{\ref*{eq:kl_grad}}
\end{align*}
With a stop-gradient teacher, the forward KL divergence splits into a cross-entropy and the
teacher's entropy:
\begin{align*}
    D_\mathrm{KL}\left(\stopgrad(q_t^{(k)}) \,\|\, p_t^{(k)}\right)
    = \sum_v q_{t,v}^{(k)} \log q_{t,v}^{(k)} - \sum_v q_{t,v}^{(k)} \log p_{t,v}^{(k)}
    = -H\left(q_t^{(k)}\right) + \mathrm{CE}\left(q_t^{(k)}, p_t^{(k)}\right).
\end{align*}
The entropy term does not depend on $\hidden_t^{(k)}$, because the teacher is held fixed. Applying
\cref{eq:app_ce_logit_grad} with $y = q_t^{(k)}$ and the same chain rule as above,
\begin{align*}
    \nabla_{\hidden_t^{(k)}} D_\mathrm{KL}\left(\stopgrad(q_t^{(k)}) \,\|\, p_t^{(k)}\right)
    = \nabla_{\hidden_t^{(k)}} \mathrm{CE}\left(q_t^{(k)}, p_t^{(k)}\right)
    = W_\theta^\intercal\left(p_t^{(k)} - q_t^{(k)}\right) = \gsoft.
\end{align*}

\subsection{From hidden states to parameters}
The decomposition in \cref{eq:hard_mtp_grad_decompose},
\begin{align*}
    g_\text{hard} = W_\theta^\intercal\left(p_t^{(k)} - q_t^{(k)} + q_t^{(k)} -x_{t+k+1}\right)
    = \underbrace{W_\theta^\intercal\left(p_t^{(k)} - q_t^{(k)}\right)}_{\gsoft} + \underbrace{W_\theta^\intercal\left(q_t^{(k)} -x_{t+k+1}\right)}_{\gres},
    \tag{\ref*{eq:hard_mtp_grad_decompose}}
\end{align*}
follows from \cref{eq:hard_mtp_grad} by adding and subtracting $q_t^{(k)}$, and holds at the MTP hidden
state. It carries over exactly to every parameter $\phi$ upstream of $\hidden_t^{(k)}$, i.e.\ in the
MTP heads and the shared trunk: for a fixed forward pass, backpropagation is linear in the gradient
arriving at $\hidden_t^{(k)}$,
\begin{align*}
    \nabla_\phi = \left(\frac{\partial \hidden_t^{(k)}}{\partial \phi}\right)^{\!\intercal} \left(\gsoft + \gres\right)
    = \left(\frac{\partial \hidden_t^{(k)}}{\partial \phi}\right)^{\!\intercal} \gsoft
    + \left(\frac{\partial \hidden_t^{(k)}}{\partial \phi}\right)^{\!\intercal} \gres.
\end{align*}
This holds for both parallel and sequential MTP heads; in the sequential case, $\hidden_t^{(k)}$
also depends on $\hidden_t^{(k-1)}$, which only changes the Jacobian, not the linearity.

\subsection{KL gradient--recovery trick}
\label{app:recovery_derivation}
We derive \cref{eq:injected_grad,eq:recovered_kl_grad}. Recall the augmented NTP hidden states
\begin{align*}
    \tilde{\hidden}_{t} = \hidden_{t}^{(0)} + \underbrace{\sum_{\substack{k=1 \\ t-k \geq 1}}^K\left[\stopgrad(\hidden_{t-k}^{(k)}) - \hidden_{t-k}^{(k)}\right]}_{=\;0}, \tag{\ref*{eq:straight-through-grad}}
\end{align*}
and the augmented NTP loss
\begin{align*}
    \tilde{\mathcal{L}}_\text{NTP} = \sum^{T-1}_{t=1}\mathrm{CE}\left(x_{t+1}, \text{softmax}\left(W_\theta \tilde{\hidden}_{t} \right) \right). \tag{\ref*{eq:augmented_ntp_loss}}
\end{align*}

\paragraph{Forward pass.} Each bracketed term in \cref{eq:straight-through-grad} is exactly zero, so
$\tilde{\hidden}_t = \hidden_t^{(0)}$ numerically, and $\tilde{\mathcal{L}}_\text{NTP}$ takes the
same value as $\mathcal{L}_\text{NTP}$.

\paragraph{Backward pass.} Let $G_t \coloneqq \nabla_{\tilde{\hidden}_t}
\mathrm{CE}\left(x_{t+1}, \mathrm{softmax}(W_\theta \tilde{\hidden}_t)\right)$. Since
$\tilde{\hidden}_t = \hidden_t^{(0)}$ numerically, \cref{eq:app_ce_logit_grad} gives
\begin{align*}
    G_t = W_\theta^\intercal\left(p_t^{(0)} - x_{t+1}\right).
\end{align*}
The stop-gradient term has zero Jacobian, so $\partial \tilde{\hidden}_t / \partial \hidden_t^{(0)} = I$
and $\partial \tilde{\hidden}_t / \partial \hidden_{t-k}^{(k)} = -I$. The NTP hidden states
therefore receive exactly the standard NTP gradient, $\nabla_{\hidden_t^{(0)}}
\tilde{\mathcal{L}}_\text{NTP} = G_t$. The MTP hidden state $\hidden_t^{(k)}$ appears only in
$\tilde{\hidden}_{t+k}$, so, using $q_t^{(k)} = p_{t+k}^{(0)}$,
\begin{align*}
    \nabla_{\hidden_t^{(k)}} \tilde{\mathcal{L}}_\text{NTP} = -G_{t+k}
    = -W_\theta^\intercal\left(p_{t+k}^{(0)} - x_{t+k+1}\right)
    = -W_\theta^\intercal\left(q_t^{(k)} - x_{t+k+1}\right) = -\gres,
\end{align*}
which is \cref{eq:injected_grad}. The NTP target at position $t+k$ exists exactly when the $k$-th MTP
target at position $t$ does ($t \le T-k-1$), so the two losses cover the same positions. Adding the
hard MTP gradient and using \cref{eq:hard_mtp_grad_decompose},
\begin{align*}
    \boxed{\nabla_{\hidden_{t}^{(k)}}\left(\tilde{\mathcal{L}}_\text{NTP} + \mathcal{L}^{(k)}_\text{MTP} \right) = g_\text{hard} - \gres= \gsoft} \tag{\ref*{eq:recovered_kl_grad}}
\end{align*}
since $g_\text{hard} - \gres = \left(\gsoft + \gres\right) - \gres = \gsoft$.

\paragraph{Output head.} The augmentation leaves the output head's gradient unchanged. For the NTP
loss,
\begin{align*}
    \nabla_{W_\theta} \mathrm{CE}\left(x_{t+1}, \mathrm{softmax}(W_\theta \tilde{\hidden}_t)\right)
    = \left(p_t^{(0)} - x_{t+1}\right) \tilde{\hidden}_t^\intercal
    = \left(p_t^{(0)} - x_{t+1}\right) \hidden_t^{(0)\intercal},
\end{align*}
and the MTP loss still contributes its cross-entropy gradient
$\left(p_t^{(k)} - x_{t+k+1}\right) \hidden_t^{(k)\intercal}$. Hence $W_\theta$ is trained with cross-entropy,
while the MTP heads and the shared trunk receive the KL gradient (\cref{tab:gradient_routing}).

\subsection{Loss weighting and per-head normalization}
\label{app:loss_weighting}

In practice, the MTP losses are weighted by $\lambda_\text{MTP}/K$ and each loss is averaged over its
own targets. Let $N_0$ be the number of NTP targets and $N_k$ the number of targets of the $k$-th MTP
head, so that
\begin{align*}
    \mathcal{L} = \frac{1}{N_0}\sum_t \mathrm{CE}\left(x_{t+1}, p_t^{(0)}\right)
    + \frac{\lambda_\text{MTP}}{K} \sum_{k=1}^K \frac{1}{N_k}\sum_t \mathrm{CE}\left(x_{t+k+1}, p_t^{(k)}\right).
\end{align*}
The MTP loss then contributes $\frac{\lambda_\text{MTP}}{K N_k}\, g_\text{hard}$ at $\hidden_t^{(k)}$,
while routing the NTP gradient as above contributes only $-\frac{1}{N_0}\,\gres$, since the NTP loss
is averaged over $N_0$ targets. Scaling each straight-through term by
\begin{align*}
    s_k = \frac{\lambda_\text{MTP}\, N_0}{K\, N_k},
    \qquad
    \tilde{\hidden}_t = \hidden_t^{(0)} + \sum_k s_k\left[\stopgrad(\hidden_{t-k}^{(k)}) - \hidden_{t-k}^{(k)}\right],
\end{align*}
leaves the forward pass unchanged and makes the routed gradient $-\frac{s_k}{N_0}\gres =
-\frac{\lambda_\text{MTP}}{K N_k}\gres$, so that
\begin{align*}
    \nabla_{\hidden_t^{(k)}} \mathcal{L}_\text{Soft-MTP}
    = \frac{\lambda_\text{MTP}}{K N_k}\left(g_\text{hard} - \gres\right)
    = \frac{\lambda_\text{MTP}}{K N_k}\, \gsoft,
\end{align*}
which is exactly the gradient of the correspondingly weighted KL objective
$\frac{\lambda_\text{MTP}}{K}\sum_k \frac{1}{N_k}\sum_t D_\mathrm{KL}\left(\stopgrad(q_t^{(k)}) \,\|\,
p_t^{(k)}\right)$. With sum rather than mean reduction, $N_0 = N_k = 1$ and $s_k = \lambda_\text{MTP}/K$.
In the implementation of \cref{alg:full_soft_mtp_code}, the labels are pre-shifted so that $N_0 = T$
and $N_k = T-k$, giving $s_k = \lambda_\text{MTP}\, T / \left(K (T-k)\right)$.

\section{Meta-style MTP results}
\label{app:meta_mtp_results}

\begin{figure}[H]
     \centering
     \begin{subfigure}[b]{0.45\textwidth}
         \centering
         \includegraphics[width=\textwidth]{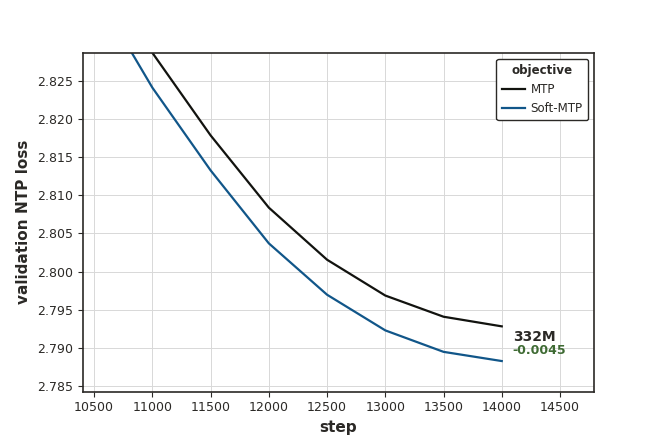}
         \caption{NTP loss.}
         \label{fig:meta_ntp_loss}
     \end{subfigure}
     \hfill
     \begin{subfigure}[b]{0.45\textwidth}
         \centering
         \includegraphics[width=\textwidth]{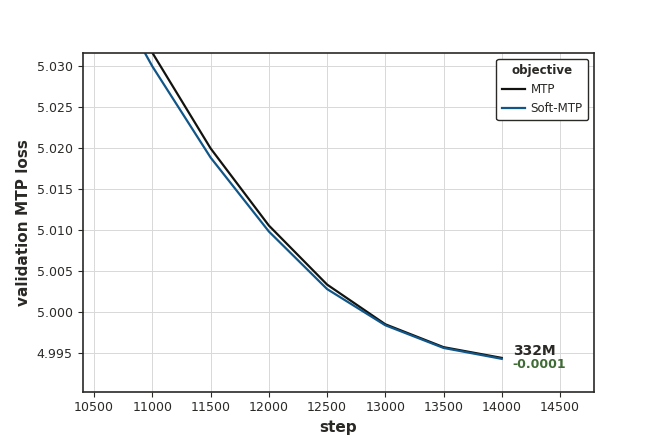}
         \caption{MTP loss.}
         \label{fig:meta_mtp_head_loss}
     \end{subfigure}
     \caption{Validation NTP and MTP losses of standard MTP and Soft-MTP with Meta-style MTP heads at 332M.}
     \label{fig:meta_mtp_loss}
\end{figure}

We pre-train standard MTP and Soft-MTP at the 332M scale using the Meta-style MTP architecture. Both runs use $K=3$ and $\lambda_\text{MTP}=1$ and train for 7.3B FineWeb-Edu tokens. \Cref{fig:meta_mtp_loss} shows that Soft-MTP attains lower NTP and MTP loss than standard MTP at the end of training.

\section{Experiment Setup and Hyperparameters}
\label{app:exp_hyperparam}

\paragraph{Architecture.} We use a LLaMA-style \citep{touvron2023llama} pre-norm Transformer with RMSNorm, RoPE, and SwiGLU activations. For experiments in the main body, MTP uses the DeepSeek MTP architecture \citep{deepseekai2024deepseekv3technicalreport}. Following the implementation of DeepSeek MTP, and confirmed across other architectures with MTP such as Nemotron 3 \citep{nvidia2025nvidianemotron3efficient}, the NTP and each MTP head have their own final RMSNorm layers.

\paragraph{Model sizes.} We fix the depth at $L=36$ layers and vary only the width, so that each one-layer MTP head is roughly a
constant fraction of the model parameters at every scale.

\begin{center}\small
\begin{tabular}{lcccc}
\toprule
 & \textbf{167M} & \textbf{332M} & \textbf{689M} & \textbf{1.2B} \\
\midrule
Total parameters          & 178.4M & 356.9M & 745.0M & 1273.0M \\
Next-token-prediction path & 167.2M & 332.1M & 689.3M & 1173.9M \\
MTP heads                 & 11.2M  & 24.8M  & 55.8M  & 99.1M \\
\bottomrule
\end{tabular}
\end{center}

\paragraph{Optimizer and schedule.} We use Muon \citep{jordan2024muon} on the 2D hidden matrices with
Newton--Schulz orthogonalization and the
update-RMS rescaling of \citet{liu2025muon}. The embeddings, the output head and all norms use AdamW
\citep{loshchilov2019decoupledweightdecayregularization}. Following the improved recipe of \citet{dao2024transformers}, we set the peak learning rate to $5\times$ the
GPT-3 value for the corresponding model size, with linear warmup and cosine decay to $10^{-5}$.

\paragraph{Code and math fine-tuning.} We follow the continued-training setup of
\citet{zuhri2026predicting}: $4.2$B tokens of OpenMathInstruct-2 \citep{toshniwal2025openmathinstruct} for math, and $21.0$B tokens of the Python subset of Stack-Edu \citep{allal2025smollm2} for code. The learning rate decays with a cosine schedule to $10\%$ of its peak, gradients are clipped at $1.0$, and code is
evaluated with pass@16, pass@32 and pass@64 on HumanEval and MBPP. We deviate from their setup by fine-tuning with the NTP loss alone, i.e., MTP heads are discarded, and by using the Muon optimizer.

\begin{table}[htbp]
  \centering
  \resizebox{\textwidth}{!}{%
  \begin{tabular}{lcccc}
    \toprule
    & \textbf{FineWeb-Edu pre-training} & \textbf{\Cref{subsection:soft_mtp_wins_hard_mtp} $K / \lambda_\text{MTP}$ sweeps} & \textbf{Math fine-tuning} & \textbf{Code fine-tuning} \\
    \midrule
    \multicolumn{5}{l}{\textbf{Data}} \\
    Corpus & FineWeb-Edu & FineWeb-Edu & OpenMathInstruct-2 & Stack-Edu (Python) \\
    Sequence length & 2048 & 2048 & 2048 & 2048 \\
    Steps & 60\,000 & 14\,000 & 8\,000 & 40\,000 \\
    Tokens & 31.5B & 7.34B & 4.19B & 21.0B \\
    \midrule
    \multicolumn{5}{l}{\textbf{Training parameters}} \\
    Batch size (sequences) & \multicolumn{4}{c}{256 \;=\; 524\,288 tokens / step} \\
    Peak learning rate & 2e-3 / 1.5e-3 / 1.25e-3 / 1e-3 & 1.5e-3 & 4e-5 & 5e-5 \\
    LR schedule & cosine & cosine & cosine & cosine \\
    LR floor & $10^{-5}$  & $10^{-5}$ & $0.1\times$ peak lr & $0.1\times$ peak lr \\
    Warmup steps & 2\,000 & 2\,000 & 800 & 2\,000 \\
    Weight decay & 0.1 & 0.1 & 0.1 & 0.1 \\
    Clip gradient norm & 1.0 & 1.0 & 1.0 & 1.0 \\
    Optimizer & \multicolumn{4}{c}{\emph{Muon} \citep{jordan2024muon}, momentum 0.95, Nesterov, 5 Newton--Schulz steps, update-RMS scaling \citep{liu2025muon}} \\
    & \multicolumn{4}{c}{\emph{AdamW} ($\beta_1=0.9,\ \beta_2=0.95$) for embeddings, output head and norms} \\
    \midrule
    \multicolumn{5}{l}{\textbf{Model parameters}} \\
    Layers $L$ & 36 & 36 & 36 & 36 \\
    Hidden dimension $d$ & 512 / 768 / 1152 / 1536 & 768 & 1536 & 1536 \\
    Attention heads & 8 / 12 / 18 / 24 & 12 & 24 & 24 \\
    Head dimension & \multicolumn{4}{c}{64} \\
    Vocabulary & \multicolumn{4}{c}{50\,304 (GPT-2 BPE, padded), untied embeddings} \\
    \midrule
    \multicolumn{5}{l}{\textbf{MTP parameters}} \\
    Architecture & DeepSeek & DeepSeek \emph{and} Meta parallel & \multicolumn{2}{c}{--- (heads discarded)} \\
    Horizon $K$ & 3 & 3--8 (3 for the weight sweep) & \multicolumn{2}{c}{---} \\
    $\lambda_{\text{MTP}}$ & 1.0 & 0.25--1.0 (1.0 for the horizon sweep) & \multicolumn{2}{c}{---} \\
    Head layers / depth & 1 & 1 & \multicolumn{2}{c}{---} \\
    \bottomrule
  \end{tabular}}
  \caption{Training, model and MTP hyperparameters. The global batch is identical everywhere
  ($256\times2048$ tokens). Both fine-tunes are trained with next-token prediction only.}
  \label{tab:exp_hyperparams}
\end{table}

\section{\Cref{subsection:effect_of_residual} Analyses}
\label{app:diagnostics}

All three diagnostics below decompose the MTP gradient at the MTP hidden state into its two
components using the exact identity of \cref{eq:hard_mtp_grad_decompose},
$g_{\text{hard}} = \gsoft + \gres$, and then ask what each component does. They are measured
\emph{offline} on frozen checkpoints. Unless stated otherwise, the estimator uses $M=256$ independent minibatches of $4$ sequences each.

\subsection{Gradient signal-to-noise ratio}
\label{app:snr}

Let $\{g_i\}_{i=1}^{M}$ be the per-minibatch mean gradients of one component, with mean $\mu$ and
covariance $\Sigma$. Writing $\bar g = \tfrac1M\sum_i g_i$,
\begin{align*}
  \text{signal} = \lVert \bar g \rVert^2, \qquad
  \text{total} = \tfrac1M\sum_i \lVert g_i \rVert^2, \qquad
  \text{noise} = \text{total} - \text{signal},
\end{align*}
and $\mathrm{SNR} = \text{signal}/\text{noise}$. Its reciprocal is the gradient noise scale of
\citet{mccandlish2018}, measured in units of minibatches. The raw ratio is biased: the measured signal
carries a $\mathrm{tr}\,\Sigma/M$ sampling floor and the measured noise is low by a factor $(M-1)/M$. We report the debiased value throughout.

\subsection{Share of the MTP gradient}
\label{app:share}

Because $g_{\text{hard}} = \gsoft + \gres$ holds exactly and per batch, each component's contribution
to the applied gradient is its projection onto it:
\begin{align*}
  \mathrm{share}_c = \frac{\langle g_{\text{hard}},\, g_c\rangle}{\lVert g_{\text{hard}}\rVert^2},
  \qquad
  \mathrm{share}_{\text{KL}} + \mathrm{share}_{\text{res}} = 1 \ \text{exactly.}
\end{align*}
Expanding $\lVert g_{\text{hard}}\rVert^2 = \lVert \gsoft\rVert^2 + \lVert \gres\rVert^2 +
2\langle \gsoft, \gres\rangle$ expresses both shares in terms of three scalars, $\lVert \gsoft\rVert^2$,
$\lVert \gres\rVert^2$ and $\langle \gsoft, \gres\rangle$, which we pool over batches before taking the
ratio. Because the shares sum to one by construction, they decompose $g_{\text{hard}}$ rather than
merely comparing magnitudes. A share can be negative when the two components point in opposite directions.

\subsection{Proxy effect of each component on the NTP loss}
\label{app:dln}

To ask what a unit step along one gradient component would do to the next-token loss, we take that step in hidden space and re-evaluate. With $H$ the (detached) trunk hidden state, $g_c$ the component's gradient backpropagated from the MTP hidden state through the MTP head to $H$, and $\mathcal{L}_\text{NTP}$ the NTP cross-entropy,
\begin{align*}
  \Delta\mathcal{L}_\text{NTP}^{(c)} \;=\; \mathcal{L}_\text{NTP}\!\left(\mathrm{RMSNorm}_\text{NTP}(H - g_c)\right)
  \;-\; \mathcal{L}_\text{NTP}\!\left(\mathrm{RMSNorm}_\text{NTP}(H)\right),
  \qquad \textbf{positive} = \textbf{harm},
\end{align*}
evaluated separately for $c \in \{\text{KL}, \text{residual}\}$; positive values mean the step increases the NTP loss.  Note that this proxy only probes what one component's direction would do to the NTP loss at this state, not what training actually does to the final model.

\end{document}